\documentclass[dvipsnames,format=sigconf]{acmart}
\usepackage{tabularray}
\UseTblrLibrary{siunitx}

\AtBeginDocument{%
  \providecommand\BibTeX{{%
    \normalfont B\kern-0.5em{\scshape i\kern-0.25em b}\kern-0.8em\TeX}}}

\copyrightyear{2023}
\acmYear{2023}
\setcopyright{rightsretained}
\acmConference[GECCO '23]{Genetic and Evolutionary Computation Conference}{July 15--19, 2023}{Lisbon, Portugal}
\acmBooktitle{Genetic and Evolutionary Computation Conference (GECCO '23), July 15--19, 2023, Lisbon, Portugal}

\begin{document}

\title{The Impact of Asynchrony on Parallel Model-Based EAs}

\author{Arthur Guijt}
\email{Arthur.Guijt@cwi.nl}
\orcid{0000-0002-0480-2129}
\affiliation{%
  \institution{Centrum Wiskunde \& Informatica}
  \city{Amsterdam}
  \country{The Netherlands}
}

\author{Dirk Thierens}
\email{D.Thierens@uu.nl}
\orcid{0000-0002-9308-5159}
\affiliation{%
  \institution{Utrecht University}
  \city{Utrecht}
  \country{The Netherlands}
}

\author{Tanja Alderliesten}
\email{T.Alderliesten@lumc.nl}
\orcid{0000-0003-4261-7511}
\affiliation{%
  \institution{Leiden University Medical Center}
  \city{Leiden}
  \country{The Netherlands}
}

\author{Peter A.N. Bosman}
\email{Peter.Bosman@cwi.nl}
\orcid{0000-0002-4186-6666}
\affiliation{%
  \institution{Centrum Wiskunde \& Informatica}
  \city{Amsterdam}
  \country{The Netherlands}
}
\affiliation{%
  \institution{Delft University of Technology}
  \city{Delft}
  \country{The Netherlands}
}


\begin{abstract}
    In a parallel EA one can strictly adhere to the generational clock, and wait for all evaluations in a generation to be done. However, this idle time limits the throughput of the algorithm and wastes computational resources. Alternatively, an EA can be made asynchronous parallel. However, EAs using classic recombination and selection operators (GAs) are known to suffer from an evaluation time bias, which also influences the performance of the approach.
    Model-Based Evolutionary Algorithms (MBEAs) are more scalable than classic GAs by virtue of capturing the structure of a problem in a model. If this model is learned through linkage learning based on the population, the learned model may also capture biases. Thus, if an asynchronous parallel MBEA is also affected by an evaluation time bias, this could result in learned models to be less suited to solving the problem, reducing performance.
    Therefore, in this work, we study the impact and presence of evaluation time biases on MBEAs in an asynchronous parallelization setting, and compare this to the biases in GAs.
    We find that a modern MBEA, GOMEA, is unaffected by evaluation time biases, while the more classical MBEA, ECGA, is affected, much like GAs are.
\end{abstract}

\ccsdesc[500]{Mathematics of computing~Evolutionary algorithms}
\ccsdesc[500]{Theory of computation~Parallel computing models}

\keywords{Genetic Algorithms, Model-Based Evolutionary Algorithms, Linkage Learning, Parallel Algorithms, Asynchronous Algorithms}

\maketitle

\section{Introduction}\label{sec:introduction}
For the optimization of many real-world problems the assessment of the quality of a solution (evaluation) involves a time-consuming process, e.g., the training of a neural network. Problems with such evaluation functions are called expensive optimization problems.

More often than not, the amount of wall time spent should be minimized. Rather than using the resources to perform these evaluations sequentially, it is preferred to use the resources simultaneously, to run evaluations in parallel. As evaluations are also commonly independent, the parallelization potential is significant, allowing us to save significant amounts of time.

However, the generational nature of an EA may limit how parallelizable the algorithm is.
EAs often follow a loop of generating offspring, evaluating them, and performing selection. Selection is only performed once all offspring solutions are evaluated, and new solutions are only generated and evaluated once selection has been performed. This describes a standard generational scheme. Alternatively, the population can be altered incrementally. For example, by generating a single solution at a time and attempting to introduce it into the population immediately after evaluation, resulting in a steady-state scheme.

In a parallel EA, unless very large population sizes are used, this limited number of solutions per generation means that processors may run out of solutions to evaluate until the next generation starts, i.e., when new offspring will be generated. Before continuing, these processors must wait on the other processors to ensure all evaluations have finished. This waiting on other processors is also called \emph{synchronization}.

\begin{figure}
    \centering
    \includegraphics[width=0.8\columnwidth]{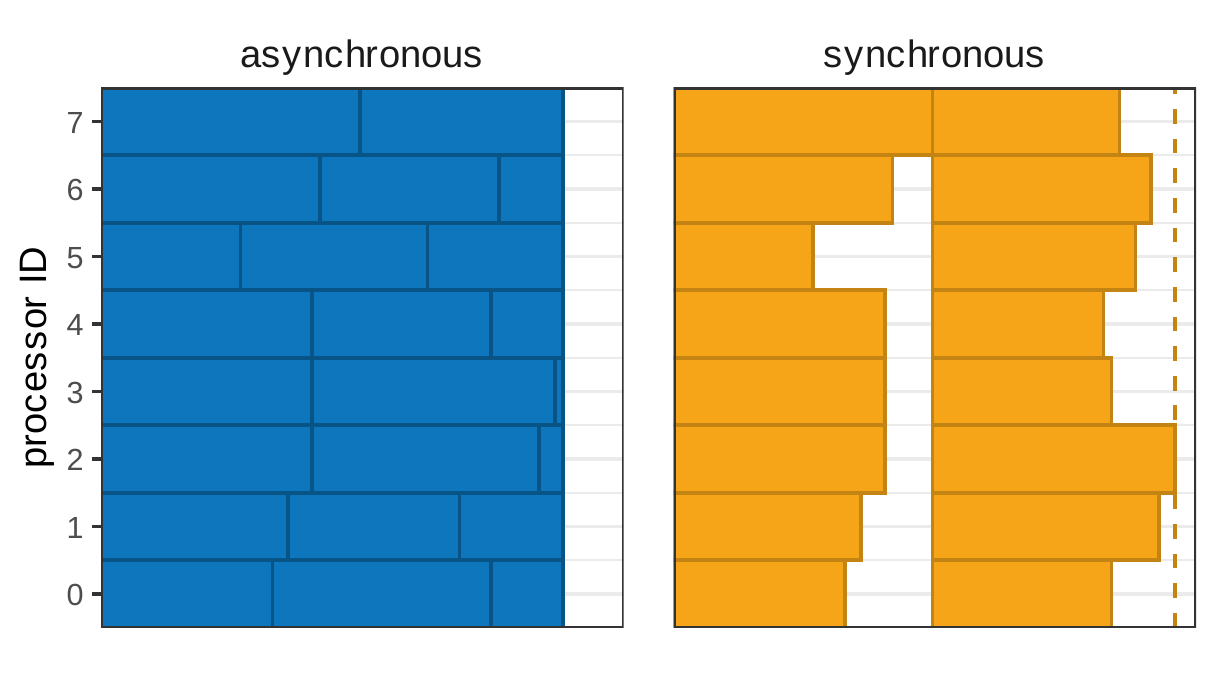}
    \vspace{-1.0em}
    \caption{Resource consumption of synchronous and asynchronous approaches under heterogeneous evaluation times. Ideally, without synchronization the target solution can be found earlier (in this example: the most expensive solution of the second 'generation').}
    \label{fig:sync-vs-async}
\end{figure}

Synchronizing is however only required if one wishes to have the exact same behavior as the sequential implementation of the EA. Alternatively, one can also asynchronously sample, evaluate, and apply selection, as is the case for the asynchronous steady-state GA~\cite{scottUnderstandingSimpleAsynchronous2015}. This may be beneficial as waiting wastes computational resources. For example, in a situation with many computing nodes, a slower node or evaluation may stop all other nodes from progressing, introducing a bottleneck into the optimization process, see Figure~\ref{fig:sync-vs-async} for an illustrative example. 
Furthermore, in the case of node or network failures, synchronization will even lock up the optimization process indefinitely.
In the remainder of this work we will refer to approaches employing synchronization as \emph{synchronous} approaches, and approaches that forgo this as \emph{asynchronous}.

However, not synchronizing does not guarantee better overall performance of the EA. For example, while in~\cite{churchillToolSequenceOptimization2013} the asynchronous configuration outperformed the synchronous configuration, in~\cite{yagoubiAsynchronousMasterSlave2012} performance degraded when using an asynchronous approach. 
Evaluation time biases were investigated in~\cite{scottUnderstandingSimpleAsynchronous2015,scottEvaluationTimeBiasAsynchronous2015,scottEvaluationTimeBiasQuasiGenerational2016}, in which was shown that there exists an evaluation time bias, i.e., the distribution of the population is biased such that it is correlated with the corresponding evaluation times, such that this bias is not explained by fitness based selection. They note that this bias depends on both the distribution of evaluation times and the number of processors. More specifically, a preference towards both short and long evaluation times on a flat fitness landscape was observed.

Even so, it is difficult to determine based on current literature, when asynchronous execution of an EA is problematic. This is in part due to how comparisons are often performed.
First, the selection procedure is often altered when switching from synchronous to asynchronous, making it impossible to distinguish effects caused by asynchrony from those caused by steady-state selection and variation.
Selection and variation are key aspects of an EA, and should also be considered as an additional influence. Furthermore, in order for effects of time biases to be interpretable, the time distributions of evaluations need to be known as well.

More generally, for EAs the population size is important as well. Different approaches may require different population sizes to perform best, especially if variation and selection are different. When switching between synchronous and asynchronous the population size should therefore be tuned again to avoid giving preferential treatment to the approach for which the population size was tuned.

Given all of this, we are particularly interested in the impact of selection and variation on the behavior of the EA. Together they induce a bias towards higher fitness solutions in the population. An oversight in how these operators work could very well induce, preserve, or halt evaluation time biases too.

In this work, we will explicitly also consider Model-based Evolutionary Algorithms (MBEAs). Through the use of linkage learning (LL) in MBEAs, variation can be performed based on inferred variable dependencies. This can result in significant performance improvements. 
To our knowledge, no prior work has studied the impact of asynchronous parallelization on MBEAs. Yet this is of interest, LL infers the structure of a problem through the use of the population. If the composition of the population is based not only on the fitness, but also the evaluation time associated with these solutions, then this will also affect the structure learning process. Therefore, while LL is known to improve performance, this could be disrupted by biases, such as evaluation time biases. 

{
\vspace{0.5em}
\noindent Our research questions are therefore:
\begin{enumerate}
\def\labelenumi{\arabic{enumi}.}
\item
  How does selection affect performance, the ability to find a solution with target fitness, under various evaluation time distributions in an asynchronous setting?
\item
  How does variation affect performance under various evaluation time distributions in an asynchronous setting?
\item
  More specifically, how are MBEAs like GOMEA and ECGA affected by the evaluation time distribution when made (a)synchronous parallel?
\end{enumerate}
}

\noindent The remainder of this work is structured as follows. First, in Section~\ref{sec:approaches} we will describe the EAs used in this work.
Following that, in Section~\ref{sec:problems}, the artificial benchmark functions and the evaluation time distributions used, in addition to a real-world NAS benchmark are described.
The remainder of experimental considerations is described in Section~\ref{sec:experimental-setup}. We discuss the results in Section~\ref{sec:results} and conclude in Section~\ref{sec:conclusion}.

\section{Approaches}\label{sec:approaches}
In this work, we include a Simple GA as described in Subsection~\ref{ssec:genetic-algorithm-ga}. To study how linkage learning (LL) and evaluation time biases interact, we use both a classic MBEA named the Extended Compact Genetic Algorithm (ECGA) and a modern MBEA named the Gene-pool Optimal Mixing Evolutionary Algorithm (GOMEA) as described in Subsections~\ref{ssec:extended-compact-genetic-algorithm-ecga} and~\ref{ssec:gene-pool-optimal-mixing-evolutionary-algorithm-gomea} respectively.

\subsection{Genetic Algorithm (GA)}\label{ssec:genetic-algorithm-ga}

In previous works, selection in GAs is often altered simultaneously with the (a)synchronous nature of the approach. For example, in~\cite{scottUnderstandingSimpleAsynchronous2015}, the synchronous configuration uses a generational selection scheme, whereas the asynchronous configuration is steady-state. A steady-state configuration exhibits different behavior compared to GAs employing a generational selection scheme \cite{syswerdaStudyReproductionGenerational1991}. We therefore also investigate  a `synchronous' steady-state variant and an asynchronous `generational' approach in addition to the original two configurations. Pseudocode for these approaches can be found in the supplementary material.

For the synchronous steady-state approach we operate in batches of $|P|$ offspring, generating each offspring at the start of the evaluation, and synchronizing until all $|P|$ offspring solutions are sampled and evaluated. Steady-state selection is then performed once the evaluation of an offspring solution is finished. For this we opt to randomly select a solution from the population and replace this solution if it is worse than the newly evaluated offspring solution. This ensures that the population's average fitness cannot decrease, and thus stops any bias contrary of improvement to fitness from taking over the population.
 When using generational selection in an asynchronous setting, every solution that completes evaluation is added into a selection pool first, similar to the approach described in \cite{scottEvaluationTimeBiasQuasiGenerational2016}. Once this pool has reached the prerequisite size, we apply the generational selection operator, replacing the entire population with the end result. We perform generational selection using a Parent + Offspring (P+O) tournament of size 4, where the number of offspring is equal to the population size. Tournaments are created based on shuffling, then by splitting the population in blocks of the tournament size, repeating as often as necessary to select P solutions.

For recombination, we use Uniform Crossover (UX) and Two-point Crossover (TPX). In addition to Subfunction Crossover (SFX) for problems for which subfunction information is available. While UX is included as a baseline, TPX and SFX are included as they suit the structure of at least one of the artificial benchmark functions described in Subsection~\ref{ssec:artificial-benchmark-functions}. Specifically, in SFX each block of variables that forms a subfunction is exchanged with $p=0.5$, perfectly mixing the blocks of the concatenated DT function, whereas TPX is especially suited for the ANKL problem due to its sequential adjacent structure. These operators should showcase how the behavior changes when a well-suited operator is used from the start and throughout the search, also providing an idealized reference for approaches employing LL.

\subsection{Extended Compact Genetic Algorithm
(ECGA)}\label{ssec:extended-compact-genetic-algorithm-ecga}

ECGA is one of the first approaches employing LL. Unlike GOMEA, explained in the next subsection, this approach still has separate recombination / variation and selection steps. This allows us to use exactly the same selection procedures as for the GA.

With ECGA, a marginal product model (MPM) is learned using a metric based on model complexity and population compression after applying selection \cite{harikLinkageLearningProbabilistic2006}. For this selection step, we apply tournament selection of size $4$. In the resulting model every variable is grouped disjoint subsets, thereby modeling each subset of variables jointly. For example, given MPM $\mathcal{M} = \{\{0, 1\}, \{2, 3\}\}$ and population (after selection) $P_M = {0011, 1100}$, allows us to sample $00$ and $11$ with, for each subset of variables, equal likelihood. Therefore, allowing us to sample $0000$, $0011$, $1100$, and $1111$.

Learning a model is costly. Therefore, performing continuous updates of this model for every solution sampled is too computationally expensive to be benchmarked properly. As such, for the asynchronous configuration the model is updated only when \(|P|\) (population size) evaluations finish. Thereby updating the model with the same frequency as the generational approach (generationally).  
While this reduced update frequency should not have too significant an impact on the learning of the MPM, solutions are sampled using out-of-date frequency estimates.

\subsection{Gene-pool Optimal Mixing Evolutionary Algorithm
(GOMEA)}\label{ssec:gene-pool-optimal-mixing-evolutionary-algorithm-gomea}

While ECGA utilizes LL, it does so differently from most modern MBEAs. Modern model-based approaches like P3 ~\cite{goldmanParameterlessPopulationPyramid2014,goldmanFastEfficientBlack2015}, DSGMA-II ~\cite{hsuOptimizationPairwiseLinkage2015,chenTwoedgeGraphicalLinkage2017}, and GOMEA~\cite{dushatskiyParameterlessGenepoolOptimal2021} all use an incremental change and accept/deny mechanism, reminiscent of local search. This allows these approaches to utilize
non-disjoint models, like the Incremental Linkage Set (ILS) in DSMGA and the Linkage Tree (LT) in GOMEA and P3. In this work we will be using the LT, which is constructed through UPGMA hierarchical clustering applied to normalized mutual information (NMI) of the variables in the population.  Each of these models are often represented as a Family of Subsets (FOS), i.e., a list of subsets containing variables to which variation should be applied jointly. For the LT the resulting tree is flattened such that each node in the tree corresponds to a subset of variables. 

Variation and selection are performed using Gene-pool Optimal Mixing (GOM). In GOM changes are made to subsets of variables as defined by a Family of Subsets, sampled from the population. After each change solutions immediately compete against their parent. Because of this, changes are evaluated more directly, preventing changes to other variables from being a source of noise for assessing the quality of the current change~\cite{dushatskiyParameterlessGenepoolOptimal2021}. If no change was made to a solution through all steps in GOM, or the strict non-improvement stretch of a solution ($1 + \left\lfloor\log_2(|P|)\right\rfloor$) was reached, Forced Improvements (FI) are applied. FI is GOM where the donor is the current best solution (elitist). If FI fails to improve a solution, the solution is replaced with the current elitist.
While this integrated variation and selection operator prevents us from changing the selection operator, this combined recombination and selection process is interesting in itself. 

\subsubsection{Synchronous} The synchronous approach closely follows the sequential version of GOMEA. Every generation starts with learning a linkage model. Then, an offspring population is made, initially containing a copy of each individual in the population. GOM and potentially FI are then scheduled to be applied in parallel on each of these offspring solutions, leaving the population from which is sampled during GOM, unchanged. When processors are available, and there are still offspring solutions left on which GOM needs to be applied, GOM is applied to of the offspring solutions in parallel. Each application of GOM is scheduled continuously until all involved evaluations have completed. Once GOM (and FI) complete the processor is freed up again. A generation ends once all solutions had GOM applied to it once.

\subsubsection{Asynchronous} When making GOMEA asynchronous, there is no longer a generational offspring population which is generationally improved. Instead, GOM is scheduled to be applied after initialization, and after completing GOM, using a queue. Once a processor has finished its current task, the next task from the queue gets executed. 
At the start of (asynchronous) GOM, if GOM has been applied $|P|$ (population size) times, a new FOS is learned. Furthermore, a copy of the population is made. This effectively turns every application of GOM into its own mini-generation.  Consequently, at the end of GOM we copy the generated offspring to the population from which is sampled. We label this configuration "a/e", for \emph{asynchronous end}. However, this configuration leads to significant use of out-of-date information: none of the accepted changes of solutions undergoing GOM are visible to other solutions. As such we also evaluate an alternative configuration that copies the offspring to the shared population at \emph{intermediate} stages during GOM, not just at the end. This configuration is labelled "a/i".

\section{Problems}\label{sec:problems}

Previous work has indicated that heritable heterogeneous evaluation times can negatively influence the behavior of an asynchronous EA~\cite{scottUnderstandingSimpleAsynchronous2015}. Heritable heterogeneous evaluation times concern variations in evaluation time associated with the genotype itself, rather than external factors like machine load. A good example of a problem with this property is training a neural network, which we will discuss in Subsection~\ref{ssec:nasbench-301}.

However, both the fitness landscape and evaluation times of such a problem are generally complex, which may complicate analysis. We will therefore first describe  benchmark functions with varying kinds of landscapes and structure, for which we will also vary the evaluation times associated with the solutions, in Subsection~\ref{ssec:artificial-benchmark-functions}.

\subsection{Artificial Benchmark
Functions}\label{ssec:artificial-benchmark-functions}
In~\cite{scottUnderstandingSimpleAsynchronous2015} it is indicated that evaluation times need to be heritable, and as such we will focus on this aspect. Similarly, evaluation times need to be preserved under variation too, as otherwise no time bias can develop. Furthermore, in~\cite{scottUnderstandingSimpleAsynchronous2015} settings in which the evaluation cost was perfectly positively and negatively correlated were investigated - and found no significant difference in performance. As any bias contrary to the improvement of fitness would be unlikely to survive selection, performance of an EA is likely not impacted. 
We therefore select a distribution based on the genotype which is not a simple function of the fitness.

Furthermore, as the results in~\cite{scottEvaluationTimeBiasAsynchronous2015} indicate a bias towards the extreme evaluation times, we will control where in our benchmark functions these extremes are located. The evaluation time setting in this work is expressed as a ratio \(a:b\), where $b$ is the cost of the optimum \(s^{*}\) and $a$ is the cost of the bitwise complement \(\bar{s^{*}}\) - i.e., the solution with all bits flipped. The evaluation time \(E(s)\) of a solution $s$, given the normalized Hamming distance \(H(s,\ s^{*})\) between this solution $s$ and the optimum $s^{*}$ is then:
\begin{align}
E(s) = H\left( s,\ s^{*} \right)\ a + \left( 1 - H\left( s,s^{*} \right) \right)\ b
\end{align}
We choose to use the ratios $100:1$, $10:1$, $2:1$, $1:1$, $1:2$, $1:10$ and $1:100$, that is, ranging from a cheaper optimum to a more expensive optimum.


\subsubsection{Concatenated Deceptive Trap
(DT)}\label{ssec:concatenated-deceptive-trap-dt}

The first problem we will use is the concatenated DT
function~\cite{debSufficientConditionsDeceptive1994, whitleyFundamentalPrinciplesDeception1991}. It consists of \(n\) blocks of size \(k\) which are concatenated together, forming a full string of length $\ell = nk$. In this work, we will only consider the case for \(k = 5\). 
In order to evaluate the function, first the number of `1' bits within the block \(b\) is counted. Following that, the \(DT\) function is applied to the unitation of each block $0$ through $n - 1$ and aggregated by summation:
\begin{align}
DT(u) &= \left\{ \ \begin{matrix}
k & u = k \\
k - u - 1 & \text{otherwise} \\
\end{matrix} \right.\\\
f(x) &= \sum_{b=0}^{n-1}{DT\left(\sum_{i=bk}^{bk+k-1}x_i\right)}
\end{align}

As there are many search paths indicating that more zeroes is better, high-fitness solutions consisting of zeroes are easiest to find. The solution consisting of all zeroes is referred to as the deceptive attractor. Yet, the needle in a haystack where \(u = k\), has better fitness: \(k\) as opposed to \(k - 1\) for the deceptive attractor. It is therefore all ones that is actually the optimum to this function. In order to solve this problem in a scalable manner, variables should be exchanged at the level of these blocks~\cite{thierensScalabilityProblemsSimple1999}, for example by recognizing block structure by using linkage learning. 


The complement of the optimum to this problem consists of all zeroes, and is the solution consisting of only attractors. We have defined the range of evaluation times depending on these two solutions. A preference towards cheaper solutions could therefore make a cheap-to-evaluate attractor even more attractive.

\subsubsection{Adjacent NK-Landscapes
(ANKL)}\label{sssec:adjacent-nk-landscapes-ankl}

While the (additively decomposable) concatenated DT function is difficult to solve with a local searcher, it is separable. This makes the problem easy to solve if the separability is known, or when this separability can be inferred. We therefore also consider the non-separable problem of ANKL. This problem consists of overlapping blocks consisting of \(k\) adjacent variables with some stride \(s\) from block to block. In this work, we consider \(k = 5\) and \(s = 2\). For each block $i$ a randomly generated function $f_i$ is defined that maps each genotype of this block to a value ranging from $[0, 1]$. Given random functions \(f_{0}\) through \(f_{n-1}\), where $x$ is a genotype of length $\ell = sn+k-1$:
\begin{align}
f(x) = \sum_{i = 0}^{n-1}{f_{i}\left( x_{si},\ldots, \ x_{si + k - 2},\ x_{si + k - 1} \right)}
\end{align}

Due to this more complex overlapping structure as well as the random nature of the subfunctions, it will be more difficult for the linkage learning approaches to configure the linkage model appropriately. As a result, such approaches require more generations to obtain a suitable model. This may therefore provide more time for any evaluation-time biases to steer the population towards or away from the optimum, potentially causing structure to be found earlier, or preventing the structure from being found at all.

\subsection{NASBench 301}\label{ssec:nasbench-301}

While benchmark functions are interesting and useful for analysis, they are not necessarily representative of practical problems. Real-world problems often contain elements that simpler benchmark functions do not  account for.

We will therefore apply the aforementioned approaches to the Neural Architecture Search benchmark NASBench 301~\cite{zelaSurrogateNASBenchmarks2022}. NASBench 301 is a benchmark for neural architecture search applied to the DARTS search space. In this search space, each network is trained from scratch. In order to make this less expensive as a benchmark, the benchmark provides both a surrogate model for the fitness and the corresponding evaluation time. This allows for an efficient simulation of a run on this problem without requiring a GPU for hours.

We use version 0.9 of the XGB surrogate model for performance and LGB model for runtime. As noisy objective functions are not a subject of research for this work, we have disabled noise for the performance model. For this experiment only the runtime model is used as an evaluation time distribution. The runtime model does not contain noise. 


\section{Experimental Setup}\label{sec:experimental-setup}
For all problems, a run using a specific population size is stopped if it has reached the target value, or if it has converged, i.e., when all genotypes in the population are identical. As there will be configurations and problem pairs for which the target value is not reached within a reasonable amount of time, for each set of problems, we will also be limiting the amount of time spent on a full bisection run. If bisection was terminated prematurely we will select the smallest successful population size found by bisection within the time limit.

Statistical tests are performed using the Mann-Whitney U-test~\cite{mannTestWhetherOne1947,fayWilcoxonMannWhitneyTtestAssumptions2010} with $p=0.05$ with Holm-Bonferroni~\cite{holmSimpleSequentiallyRejective1979} correction where applicable. All experiments are carried out on a machine with two AMD EPYC 7282 16-Core Processors \@ 2.8GHz, with 252~GB of RAM. Source code will be made available.

\subsection{Artificial Benchmark Functions}
Prior work~\cite{scottUnderstandingSimpleAsynchronous2015,scottEvaluationTimeBiasAsynchronous2015,scottEvaluationTimeBiasQuasiGenerational2016} investigated the distribution of evaluation times. However, if an approach is influenced such that the distribution of evaluation times is biased, this does necessarily indicate a negative impact on the performance of an approach, like the time required to reach the optimum. We will focus on performance under various evaluation time configurations.

However, using standard performance metrics is problematic. When using the number of evaluations required to reach a target fitness, synchronous approaches are favored as waiting does not incur any penalty, whereas utilizing these resources with additional evaluations does count as additional evaluations. For our first experiment, we will be changing the distribution of evaluation times, and compare all investigated distributions. Using the wall time in this comparison is problematic as the total wall time spent will change with the distribution. For example, scaling all evaluation times by $10$ will not change the evaluation times with respect to one another. As such a run will proceed identically, except with evaluation times that are $10$ times larger. For more complex distributions it would be hard to say whether the approach is actually negatively affected by the change in evaluation times, or whether the solutions to be evaluated are more costly.

We instead choose a measure that stays constant if behavior is the same, yet still indicates when performance worsens. For this we use the minimally required population size to reach a target value as determined by bisection. This measure works for a convergent EA as the population acts as a buffer for diversity loss, the larger the population the longer it takes for biases to take over the population. If a bias leads to a decrease in diversity such that the target fitness is no longer reached, increasing the population size will likely allow for solving the problem again (assuming no limits are hit).

Furthermore, we will perform these first experiments with the number of processors equal to \(|P|\) (the population size) as this maximizes the degree of parallelization proportional to the number of evaluations performed. Additionally, given this setting, the evaluations with a fast evaluation time will also be the first to complete, which is not guaranteed with low degrees of parallelization.

The minimally required population size cannot be readily compared across different approaches for performance. 
Even if an approach has a smaller minimally required population size it may still perform more evaluations and spend more time finding the optimum than another approach. Furthermore, the number of times offspring are generated also impacts the amount of used resources. Yet as before, measuring this invariant to the evaluation time distribution is difficult. Instead, we compare how the minimally required population size scales across varying evaluation time distributions.

For these experiments, each configuration will be evaluated for 100 random seeds, with each bisection run limited to 1 hour. This is orders of magnitude more than what most approaches need, and ensures even the worst performing algorithms have a chance.

\subsection{NASBench 301}
While comparing how approaches scale across different time distributions will allow us to study the impact of evaluation time biases, a practical problem often only has a single evaluation time distribution associated with it. Furthermore, the amount of computational hardware available is not unbounded in practice. If this were not the case, one could evaluate the entire search space in parallel at the cost of the most expensive solution. A more practical goal for parallelization for a specific problem is to minimize the amount of wall time spent to hit a target fitness given a limited amount of resources. As we only regard a single evaluation time distribution, aforementioned concerns do not apply to the NASBench 301 experiment.

Therefore, for the NASBench 301 experiment we will run experiments for the simulated wall time required to reach a target value. Here, first a range bounding this minimal evaluation time is determined, followed by a modified golden section search~\cite{kieferSequentialMinimaxSearch1953}, detailed further in the supplementary material. This is to find the population size that minimizes the corresponding wall time to ensure that each approach can be compared fairly.

For this experiment the number of processors is restricted to 64. Each configuration is evaluated for 20 different random seeds for at most 16 hours, due to the more costly nature of using the surrogate over a benchmark function. As in preliminary experiments this time limit was found to be insufficient due to wasting a significant amount of time on small population sizes, we additionally require an improvement to be found every \(2e + 10P\) evaluations, where \(e\) is the number of evaluations issued at the last improvement and \(P\) is the population size.

\section{Results and Discussion}\label{sec:results}
\begin{figure*}[tbp]
    \centering
    \vspace{-0.5em}
    \includegraphics[width=1.0\textwidth]{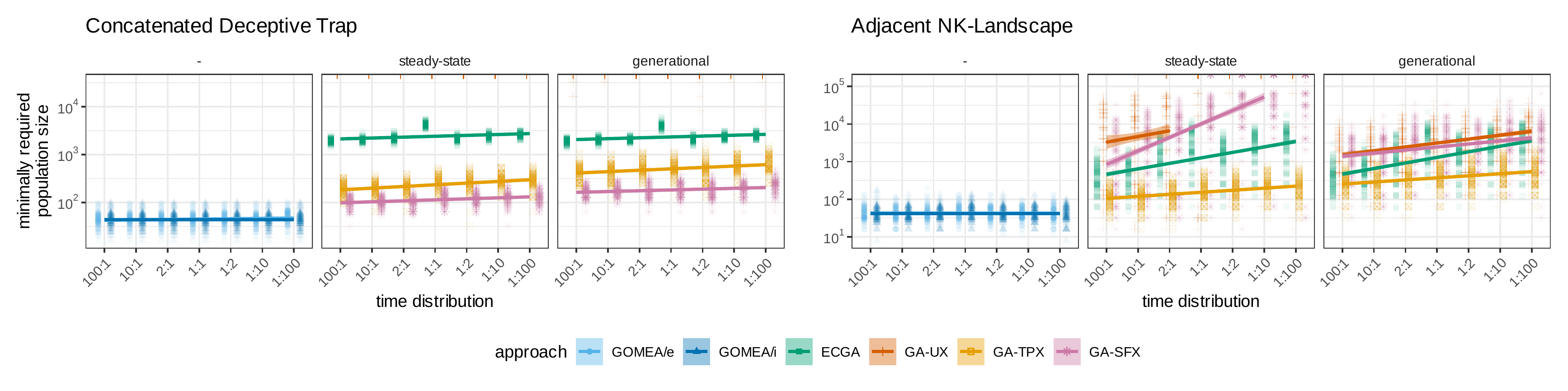}
    \vspace{-2em}
    \caption{Correlation between time distribution and minimally required population size, with to the left the optimum being cheaper and to the right the optimum being more expensive than its complement. All runs are plotted as a point with opacity. Only asynchronous configurations are shown. For the creation of the regression lines we assume that failed runs used the maximum population size tested and are not drawn if more than half of the runs did not find the optimum. (Left: Concatenated DT $l=50$, $k=5$, Right: ANKL $l=40$, $s=2$, $k=5$)}
    \label{fig:async-regression}
\end{figure*}

\begin{table*}[tbp]
    \centering
    \vspace{-0.5em}
    \caption{Median minimally required population size on DT for $\ell=50$, $k=5$. Extended table in supplementary material.}
    \label{tab:table-dt}
    \includegraphics{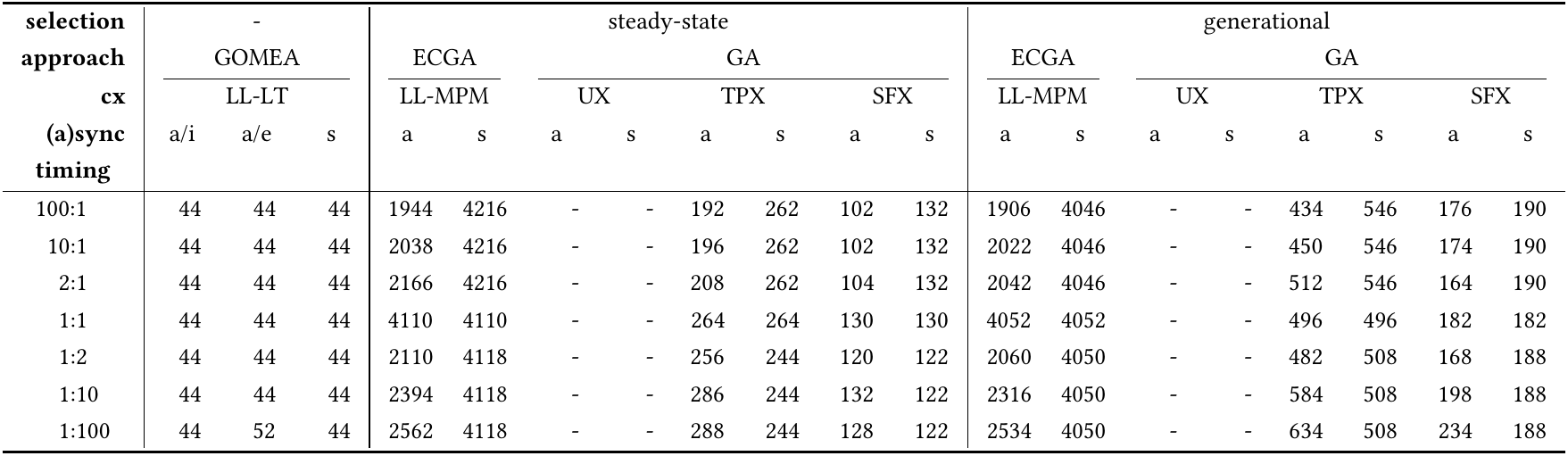}
\end{table*}
\begin{table*}[tbp]
    \centering
    \vspace{-0.5em}
    \caption{Median minimally required population size for ANKL for $\ell=40$, $s=2$, $k=5$. Extended table in supplementary material.}
    \label{tab:table-ankl}
    \includegraphics{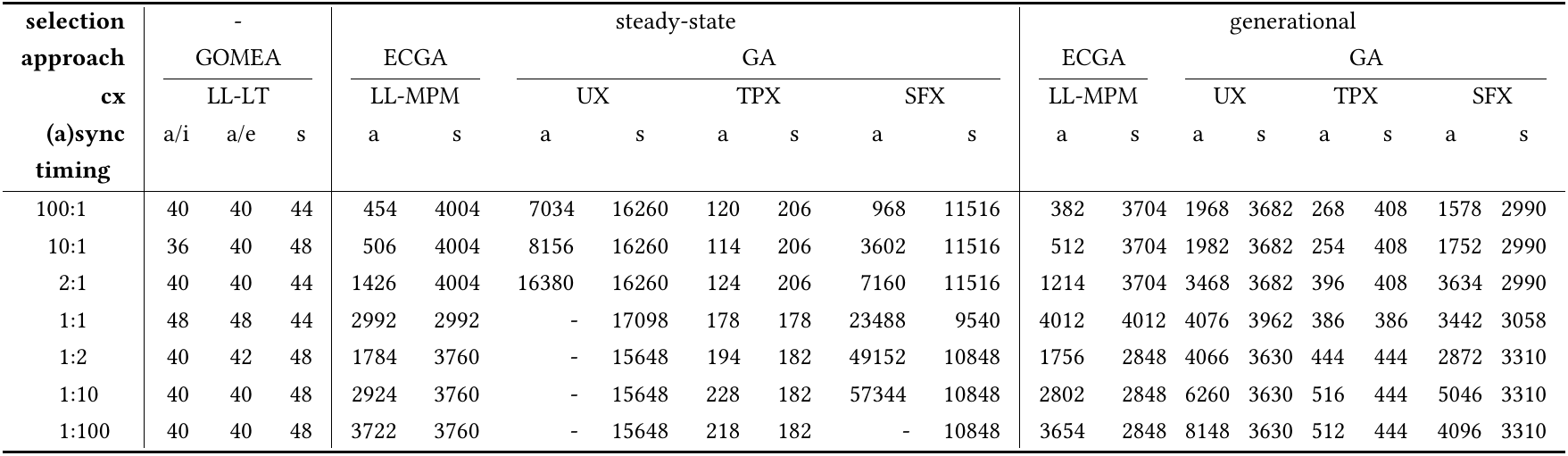}
\end{table*}
\begin{table*}[tbp]
  \centering
  \vspace{-0.5em}
  \caption{Median of minimally required amount of time to find a solution with an accuracy of 95.2727 or higher on NASBench 301 and corresponding median population size. Sample count is even, median is midpoint average.}
  \label{tab:table-nasbench}
  \includegraphics{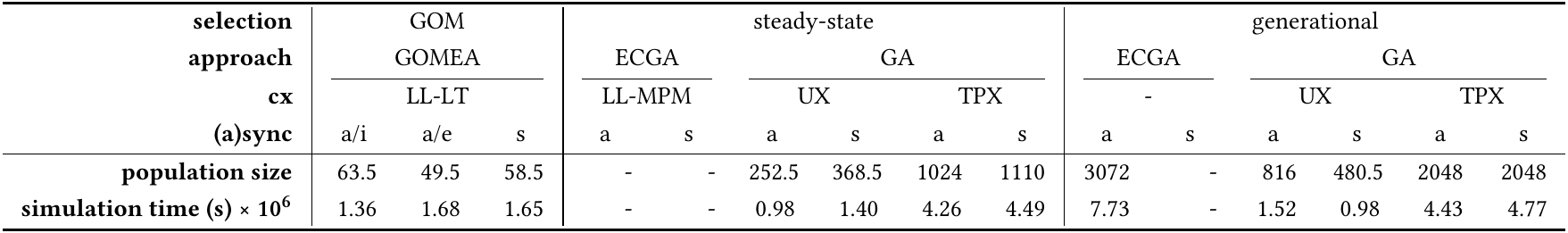}
\end{table*}

First, we will discuss the results on the artificial benchmark functions (Table~\ref{tab:table-dt} for DT and Table~\ref{tab:table-ankl} for ANKL). After this, we will discuss the results on
NASBench 301.

\subsection{Artificial Benchmark Functions} From Figure~\ref{fig:async-regression} it is apparent that asynchronous configurations experience an evaluation time bias that leads to a change in behavior. Specifically, the required population size is lower when the optimum is cheaper, i.e., is faster to evaluate, and larger when the optimum is more expensive, i.e., takes longer to evaluate. Simultaneously, synchronous approaches are invariant to the distribution of evaluation times investigated. This is in line with what would be expected based on the results in literature~\cite{scottUnderstandingSimpleAsynchronous2015,scottEvaluationTimeBiasAsynchronous2015,scottEvaluationTimeBiasQuasiGenerational2016}.

However, the extent of the differences in minimally required population size is highly dependent on the crossover used. When the most suitable crossover is used, i.e., SFX for DT and TPX for ANKL, the differences between the timing settings are small. At the same time, when an unsuitable crossover is used, differences in required population size range \emph{across orders of magnitude}. In the worst case, the problem is not solved within the allotted time for more expensive optima, as such evaluation time biases can negatively impact the performance of an approach.

The selection method can also impact how the approach scales across different evaluation time settings.
For example, on ANKL the asynchronous steady-state GAs with UX and SFX degrade substantially, even to the point of failure, whereas the same GA, but with a (pseudo-)generational scheme degrades substantially less, by at most a factor 5. Yet, with the right crossover, steady-state actually has a smaller minimally required population size, and degradation between both configurations is at most a factor of 2.

We conjecture that this is caused by the following. In a steady-state approach the solution is immediately integrated into the population.
Additionally, it is also immediately available to be used by crossover to generate new offspring. These offspring are now more likely to be fast evaluating solutions, depending on how well variation will preserve the evaluation time of a solution. This effect can accumulate over time, leading to premature convergence.

On the other hand, in a generational scheme these solutions are not immediately integrated into the population. Newly generated offspring are hence distributed according to the original distribution of solutions -- with less evaluation time bias affecting them.
Simultaneously, solutions with longer evaluation times will take longer for a processor to evaluate. During this time, no other solutions will be sampled for evaluation on this processor. In effect, the pool of currently evaluating solutions performs selection against short evaluation times. This will cause the amount of resources allotted to fast evaluating solutions to be lower. Finally, this results in less accumulation of evaluation time bias towards fast evaluating solutions, avoiding premature convergence towards such solutions.

Variation also has a notable impact on the impact of evaluation times. When variation aligns with the problem's structure, the different selection schemes unexpectedly scale similarly across evaluation time settings.
Picking the right crossover operator will help with avoiding significant degradation for asynchronous approaches, even if the approach is in a steady-state configuration.

These results are promising for MBEAs because in MBEAs problem structure is automatically learned to inform its variation operator. One could therefore expect performance to always be reasonable, resulting in the precise selection scheme mattering less.

Before we discuss the results for ECGA, we repeat that ECGA's steady-state configuration is not truly steady-state, but only applies steady-state like selection. This is because ECGA only updates its model once \(|P|\) (population size) solutions finish evaluating, as stated in
Section~\ref{ssec:extended-compact-genetic-algorithm-ecga}. This results in an approach with behavior more closely resembling that of the generational approach. The differences between the two selection methods is negligible if the distribution updates less frequently, as is the case for ECGA.

First of all, there is the oddity that the constant time distribution requires a larger population size in order to find the optimum than the other asynchronous configurations for ECGA. The required population size is more in line with the synchronous configuration than the other time distributions. We explain this as follows. In an asynchronous setting with heterogeneous evaluation times, after the first few solutions finish evaluation, not enough solutions have finished evaluation yet to update the model. As such, the next solutions to be evaluated are still sampled from the initial random model.
These solutions are therefore additional random solutions. Conversely, in the case where the evaluation time distribution is constant, all evaluations finish at the same time. This results in the model being updated. Therefore, the next solutions to be evaluated on the freed up processors, will be sampled using an \emph{updated} model, potentially with some (evaluation time) bias.

After a model update, only the solutions that are currently being evaluated can originate from an older model. There are at most "number of processors" such solutions.
In the case of the experiments above, this would be equal to the population size. This is reflected in the results: the minimally required population size is approximately twice as big, only for the constant evaluation time. The set of solutions that are currently evaluating may therefore act like an extension of the population itself.

When observing how ECGA's minimally required population size scales for different evaluation time settings in Figure~\ref{fig:async-regression}, note that it scales worse compared to a GA with a suitable crossover for ANKL, and even with a non-suitable crossover with generational selection. Variation is more than the subsets of variables that are captured in the MPM model. How the model is used is just as important. In this case, the global sampling for each subset seems to be worse than the recombinative crossover of a GA. In conclusion, though not necessarily the fault of LL, LL within an MBEA is no guarantee that variation will perform well. 

In contrast to ECGA and the GAs, GOMEA is found to be invariant to the evaluation time setting in most cases. There exists only a single time setting and problem combination for GOMEA for which the medians are statistically significantly different from the rest: 'a/e' on DT for 1:100, see Table~\ref{tab:table-dt}.
This is likely due to the combined variation and selection method used in GOMEA. As changes are made to individuals, they only compete against their parent. When parent and offspring are similar in evaluation time, this automatically results in niching behavior with respect to the evaluation times of solutions. In comparison to global selection, this approach stops solutions that evaluate quicker from taking over the population, in effect removing a large source of evaluation time bias.

\begin{figure}[bt] 
    \centering
    \includegraphics[width=0.8\columnwidth]{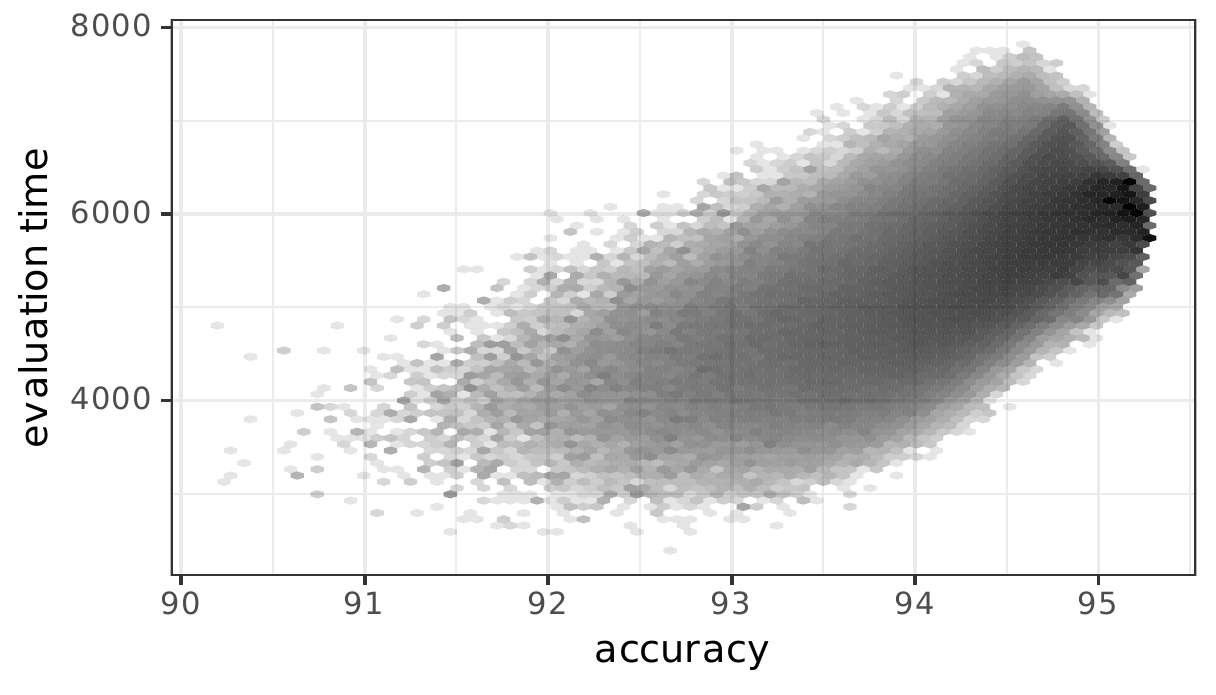}
    \caption{Objective (accuracy of the trained network) versus evaluation time sampled by runs of the approaches for NASBench 301. Showing the experienced correlation between the two for the approaches.}
    \label{fig:scatter-eval-time-acc-nasbench}
\end{figure}

\subsection{NASBench 301} For NASBench 301 the distribution of the objective and evaluation time of a solution is shown in Figure~\ref{fig:scatter-eval-time-acc-nasbench}. From the positive correlation and observations on the benchmarks one would expect to see asynchronous configurations to be outperformed by their corresponding synchronous approaches. Yet, if one refers to Table~\ref{tab:table-nasbench}, this is not the case. As a matter of fact, some of the best performing approaches are asynchronous.
We explain this as follows.

Referring to Figure~\ref{fig:scatter-eval-time-acc-nasbench}, one may note that the correlation is not perfect. The target solution for this problem is not actually the most expensive or least expensive solution, as was the case for the artificial benchmark functions.
Additionally, an EA does not utilize all of these solutions simultaneously. If we look at the correlation from a particular fitness value onwards, i.e., truncating the population, the correlation decreases as this fitness threshold increases.
Eventually, this correlation even becomes slightly negative for these data points: a setting which would actually be considered beneficial for steady-state asynchronous GAs based on our previous results.

For ECGA this is particularly notable. The approach itself is known to have trouble working with certain multi-modal functions~\cite{chuangMultivariateMultimodelApproach2010}, and it seems NASBench 301 is among these problems. Even so, the asynchronous approach has runs in which a solution with at least the target fitness was found, whereas the synchronous approach does not. This could be due to evaluation time biases helping the approach, much like what happens with the asynchronous steady-state GA for ANKL. It is therefore plausible that evaluation time biases may in part be responsible for the improvements in performance observed for asynchronous approaches, rather than only the improvements in throughput.

Furthermore, we have observed ECGA to prematurely merge FOS elements together on NASBench 301. As the model is unlikely to merge variables if they are not correlated, such a merge seems to only further reinforce correlation for this problem. Combined with the high selection pressure, this is likely to result in premature convergence to a single mode. In contrast, the linkage tree used in GOMEA does not suffer from this issue. The LT FOS always includes the univariate FOS elements: subsets with each variable on their own. Combined with the niching behavior described above, this significantly reduces the possibility of premature convergence no matter the source.

This further reinforces that an MBEA not only has to learn the right linkage from the population, but also use it in the right manner. 

For GOMEA we would expect a difference between the time required for the synchronous and asynchronous approach. Since with GOM many similar variation steps are done to a single solution sequentially, the variance in evaluation times is potentially amplified. Furthermore, population sizes are relatively small compared to the GA and ECGA. We would therefore expect the throughput for asynchronous GOMEA to be considerably higher, and as such the time required to be lower.

However, no statistically significant difference in time between the \emph{a}synchronous/e and \emph{s}ynchronous approach is observed ($p=0.52$, $U=224$, $20$ samples). Furthermore, a more detailed investigation of a single run does indicate that the amount of time spent idle for the synchronous approach is notable, on average $140059$s per processor over the entire run. In contrast, there is a statistically significant difference for the time required between the asynchronous/i and synchronous approach ($p=0.007 < 0.05$, $U=100$, $20$ samples). Effectively, when using outdated parents, offspring are more likely to have a lower fitness value. This degrades the performance of the approach, counteracting any gains in throughput.

For an asynchronous MBEA to actually gain performance compared to its synchronous counterpart, model updates should not be too infrequent and material with which is recombined should be recent. GOMEA could still be improved in this regard: each run of GOM keeps a copy of the population for sampling. As this population was created at the start of GOM, prior to the first evaluation, this population will gradually become outdated. Updating this population during GOM could result in recombination with more up-to-date solutions, potentially further improving performance.


\section{Conclusions}\label{sec:conclusion}

Answering RQ 1, we have observed that steady-state asynchronous EAs are much more vulnerable to the biases induced by heterogeneous evaluation times, compared to asynchronous EAs using a generational scheme.
Furthermore, answering RQ 2, when paired with a variation operator that is not competent in terms of improving fitness, the impact on an algorithm's capability to solve a problem can be severe. In contrast, when using well suited variation and selection operators, any differences between synchronous and asynchronous configurations become much smaller. 

For the first time we have investigated the impact of heterogeneous evaluation times on parallel MBEAs and linkage learning (LL).
The addition of LL promises to automatically align a variation operator with the structure of a problem. However, answering RQ 3, there are significant differences in the impact of evaluation times on the performance of ECGA, and GOMEA. While LL is not affected to the extent that evaluation time biases prevent the approach from finding high quality solutions, its performance greatly depends on how variation and selection are performed.

Finally, rather than negatively impacting the results, LL, when paired with the right variation and selection scheme can be a useful tool for obtaining good performance in general, even when an EA is asynchronous. GOMEA is an example of such an EA. GOMEA gets selection and variation right for all the problems evaluated, and is invariant to the choice between a synchronous or asynchronous configuration.

\begin{acks}
This publication is part of the project "DAEDALUS - Distributed and Automated Evolutionary Deep Architecture Learning with Unprecedented Scalability" with project number \grantnum{NWO}{18373} of the research programme \grantnum{NWO}{Open Technology Programme} which is (partly) financed by the \grantsponsor{NWO}{Dutch Research Council (NWO)}{}. Other financial contributions as part of this project have been provided by 
\grantsponsor{CElekta}{Elekta AB}{}
 and 
\grantsponsor{CORTECLC}{Ortec Logiqcare B.V.}{}.
\end{acks}

\bibliographystyle{ACM-Reference-Format}
\bibliography{bibliography.bib}

\end{document}


\maketitle
\tableofcontents

\section{Parallel EAs}
In the main paper we roughly describe the setup of our parallel EAs. This section serves to clarify this setup. Within our codebase we simulate the running of the algorithm such that only function evaluations cost time. Each of the evaluations is performed on one of the processors who perform work according to a queue $\mathcal{Q}$ to which the algorithms add work items. Note that work items are not necessarily a single function evaluation - in the case of GOMEA this constitutes an entire application of GOM. Once a work item is finished a new item is removed from the top of queue - or if empty, the worker waits.

A key difference between the synchronous and asynchronous approaches is how this queue is used. In the synchronous approaches the queue is used as an inner loop for which tasks are appended at the beginning of the generation, and waits until all tasks run until completion. Whereas in the asynchronous approach each task schedules a similar task to run at the end.


\begin{algorithm}
    \SetKwFunction{FInit}{Initialize}
    \SetKwFunction{FStepGeneration}{StepGeneration}
    \SetKwFunction{FAppend}{Append}
    \SetKwFunction{FSynchronize}{WaitUntilAllTasksDone}
    \SetKwFunction{FTask}{PerformTask}
    \SetKwFunction{FTaskGA}{PerformTask-GA}
    \SetKwFunction{FSampleRandomly}{SampleRandom}
    \SetKwFunction{FSampleFromP}{SampleFromPopulation}
    \SetKwFunction{FCrossover}{Crossover}
    \SetKwFunction{FEvaluate}{Evaluate}
    \SetKwFunction{FSelect}{Select}
    \SetKwFunction{FRunSync}{RunSynchronous}

    \Fn{\FInit{}}{

    \ForEach{$p \in \mathcal{P}$}{
        \Comment{\FTask is not called immediately}
        $\mathcal{Q} \gets$ \FAppend{$\mathcal{Q}$, $\lambda: $\FTaskGA{true}}\;
    }
    \FSynchronize{}
    }

    \Fn{\FStepGeneration{}}{

    \ForEach{$p \in \mathcal{P}$}{
        \Comment{\FTask is not called immediately. $\lambda$ creates a function that takes no arguments and calls the function given the arguments listed.}
        $\mathcal{Q} \gets$ \FAppend{$\mathcal{Q}$, $\lambda: $\FTaskGA{false}}\;
    }
    \FSynchronize{}
    }

    \Fn{\FRunSync{}}{
        \Comment{Termination happens as condition is met.}
        \FInit{}\;
        \While{true}{
            \FStepGeneration{}\;
        }
    }

    \Fn{\FTaskGA{is-init}}{
        \eIf{is-init}{
            $s \gets $\FSampleRandomly{}\;
            $i \gets |\mathcal{P}|$\;
            $\mathcal{P} \gets$ \FAppend{$\mathcal{P}$, $s$}\;
            $s \gets $\FEvaluate{$s$}\;
            \If{$s > \mathcal{P}[i]$}{
                \Comment{In an asynchronous approach the solution may have been replaced - this allows us to backtrack the change if the initial solution turned out to be better.}
                $\mathcal{P}[i] \gets s$\;
            }
        }{
            $p_0, p_1 \gets $\FSampleFromP{}\;
            $s \gets $\FCrossover{$p_0$, $p_1$}\;
            $s \gets $\FEvaluate{$s$}\;
            $\mathcal{P} \gets $\FSelect{$\mathcal{P}$, $s$}\;
        }
    }
    \caption{Outline of a Parallel Synchronous GA}
\end{algorithm}

\begin{algorithm}
    \SetKwFunction{FTaskGA}{PerformTask-GA}
    \SetKwFunction{FTaskSGA}{PerformTask-Async-GA}
    \SetKwFunction{FRunAsynchonous}{RunAsynchronous}
    \SetKwFunction{FWaitUntilTermination}{WaitUntilTermination}
    \SetKwFunction{FAppend}{Append}
    \Fn{\FTaskSGA{is-init}}{
        \Comment{Same actions are performed as in the synchronous version.}
        \FTaskGA{is-init}\;
        \Comment{Except, next task is queued asynchronously.}
        $\mathcal{Q} \gets$ \FAppend{$\mathcal{Q}$, $\lambda: $\FTaskSGA{false}}\;
    }
    \Fn{\FRunAsynchonous{}}{
        \ForEach{$p \in \mathcal{P}$}{
            \Comment{\FTask is not called immediately}
            $\mathcal{Q} \gets$ \FAppend{$\mathcal{Q}$, $\lambda: $\FTaskSGA{true}}\;
        }
        \Comment{Now, wait until the termination criterion is hit - tasks schedule themselves.}
        \FWaitUntilTermination{}\;
    }

    \caption{Outline of a Parallel Asynchronous GA}
\end{algorithm}

\begin{algorithm}
    \SetKwFunction{FSelectGenerationalPooling}{SelectGenerationalPooling}
    \SetKwFunction{FSelectGenerational}{SelectGenerational}
    \SetKwFunction{FSelect}{Append}
    \SetKwFunction{FAppend}{Append}
    \Fn{\FSelectGenerationalPooling{$\mathcal{P}$, $s$}}{
        \Comment{This method has additional attached data $\mathcal{O}$ - an offspring population - that is shared between invocations}
        $\mathcal{O} \gets $\FAppend{$\mathcal{O}$, $s$}\;
        \If{$|\mathcal{O}| = |\mathcal{P}|$}{
            \Comment{Enough offspring collected to apply a generational selection operator, like tournament selection.}
            $\mathcal{P} \gets$ \FSelectGenerational{$\mathcal{P}$, $\mathcal{O}$}\;
        }
    }
    \caption{Special cases for operators of a Parallel GA}
\end{algorithm}

\begin{algorithm}
    \SetKwFunction{FTaskECGA}{PerformTask-ECGA}
    \SetKwFunction{FLearnModel}{LearnModel}
    \SetKwFunction{FSample}{Sample}
    \Fn{\FTaskECGA{is-init}}{
        \If{is-init}
        {
            \Comment{Same as GA}
            ...
        }
        
        \Comment{Check if the sampling model needs to be updated. Both uses-left and $\mathcal{M}$ are shared across tasks.}
        \If{uses-left $= 0$}{
            $\mathcal{M} \gets$ \FLearnModel{$\mathcal{P}$}\;
            \Comment{This parameter setting will effectively learn the model at the start of each generation in a synchronous approach, yet also work when transferred to an asynchronous setting.}
            uses-left $\gets |\mathcal{P}|$\;
        }
        $s \gets$ \FSample{$\mathcal{M}$}\;
        uses-left $\gets$ uses-left $-1$\;
        $s \gets $\FEvaluate{$s$}\;
        $\mathcal{P} \gets $\FSelect{$\mathcal{P}$, $s$}\;
    }
    \Comment{General framework is equivalent to the GA, except the task above is used instead.}

    \caption{Task for ECGA}
\end{algorithm}

\begin{algorithm}
    \SetKwFunction{FTaskGOMEA}{PerformTask-GOMEA}
    \SetKwFunction{FLearnFOS}{LearnFOS}
    \SetKwFunction{FSampleFromP}{SampleFromPopulation}
    \SetKwFunction{FGOMFI}{GOM+FI}
    \SetKwFunction{FGOM}{GOM}
    \Fn{\FTaskGOMEA{is-init, idx}}{
        \If{is-init}
        {
            \Comment{Same as GA}
            ...\\
            \Return{}
        }
        
        \Comment{Check if the FOS needs to be updated. Both uses-left and $\mathcal{M}$ are shared across tasks.}
        \If{uses-left $= 0$}{
            $\mathcal{F} \gets$ \FLearnFOS{$\mathcal{P}$}\;
            \Comment{Much like the model used for ECGA, this will result in a new model being learned at the start of every generation in a synchronous approach.}
            uses-left $\gets |\mathcal{P}|$\;
        }
        uses-left $\gets$ uses-left $-1$\;
        \Comment{Create a copy of the population}
        $\mathcal{P}^\prime \gets \mathcal{P}$\;
        $s \gets \mathcal{P}[idx]$\;
        $s \gets$ \FGOMFI{$s$, $idx$, $\mathcal{F}$, $\mathcal{P}^\prime$}\;
        \Comment{Solution is always updated after GOM for asynchronous. For synchronous the copy back happens generationally.}
        \If{async}{
            $\mathcal{P}[idx] \gets s$\;
        }
    }

    \Fn{\FGOM{s, idx, $\mathcal{F}$, $\mathcal{P}^\prime$}}{
        \Comment{Statistics tracking for FI has been omitted.}
        \ForEach{$v \in \mathcal{F}$}{
            $s_b \gets s$\;
            $d \gets$ \FSampleFromP{$\mathcal{P}^\prime$}\;
            $s[v] \gets d[v]$\;
            \eIf{$f(s) \geq f(s_b)$}{
                $s_b \gets s$\;
                \Comment{Used in GOMEA a/i}
                \If{update-immidiately}{
                    $\mathcal{P}[idx] \gets s$\;
                }
            }{
                $s \gets s_b$\;
            }
        }
    }

    \caption{Task for GOMEA}
\end{algorithm}

\section{Modified Golden-Section Search for Population Size for the NASBench 301 experiment}
Finding the minimum time required to find the optimum is a roughly unimodal minimization problem. Too small a population size may require additional generations to find the optimum, or may not use all the resources available to a parallel GA. While a larger population requires more work per generation, which could slow down the algorithm too.

A key problem is that for some population sizes the optimum is not found at all, given our termination criteria. For small population sizes the approach is likely to prematurely converge, as such, first the bounds need to be determined.

This search starts off with finding a population size \(P_{1}\) for which the problem is solved, followed by evaluating the time necessary for a population twice the size \(P_{2} = 2P_{1}\). If the time required to solve the problem at this population size is larger, we can immediately continue with the triplet \((P_{0} = \frac{P_{1}}{2},P_{1},P_{2})\). Otherwise, we continue doubling \(P_{1}\) until this is the case. Given the triplet \((P_{0},\ P_{1},\ P_{2})\) of population sizes, we sample \(P_{3}\) at \(\frac{1}{3}\) or \(\frac{2}{3}\) between \(P_{0}\) and \(P_{2}\), such that it lies in the longest segment between \((P_{0},\ P_{1})\) or \(\left( P_{1},\ P_{2} \right)\), respectively. If the time required for \(P_{3}\) is greater than \(P_{1}\), we continue with the triplet \((P_{0},\ P_{1},\ P_{3})\), otherwise we continue with \((P_{1},P_{3},\ P_{2})\). The algorithm terminates with \(P_{1}\ \)if there are no new unevaluated population sizes between \(P_{0}\) and \(P_{2}\).

    
\section{Extended Tables}
As only the median on its own is not always as insightful, we have included additional tables containing the 25 and 75 percentiles as well. See Table~\ref{tab:table-dt-ext} for Deceptive Trap, Table~\ref{tab:table-ankl-ext} for Adjacent NK-Landscapes, and Table~\ref{tab:table-nasbench-ext} for NASBench 301.

\newgeometry{left=0.1cm,right=0.1cm,top=0.1cm,bottom=0.1cm}
\begin{landscape}
        \centering
    
\small
\begin{tblr}{
    cell{1}{2} = {c=9}{},
    cell{2}{2} = {c=9}{},
    cell{3}{2} = {c=9}{},
    cell{4}{2} = {c=3}{},
    cell{4}{5} = {c=3}{},
    cell{4}{8} = {c=3}{},
    cell{7}{2} = {r},
    cell{7}{3} = {r},
    cell{7}{4} = {r},
    cell{7}{5} = {r},
    cell{7}{6} = {r},
    cell{7}{7} = {r},
    cell{7}{8} = {r},
    cell{7}{9} = {r},
    cell{7}{10} = {r},
    cell{8}{2} = {r},
    cell{8}{3} = {r},
    cell{8}{4} = {r},
    cell{8}{5} = {r},
    cell{8}{6} = {r},
    cell{8}{7} = {r},
    cell{8}{8} = {r},
    cell{8}{9} = {r},
    cell{8}{10} = {r},
    cell{9}{2} = {r},
    cell{9}{3} = {r},
    cell{9}{4} = {r},
    cell{9}{5} = {r},
    cell{9}{6} = {r},
    cell{9}{7} = {r},
    cell{9}{8} = {r},
    cell{9}{9} = {r},
    cell{9}{10} = {r},
    cell{10}{2} = {r},
    cell{10}{3} = {r},
    cell{10}{4} = {r},
    cell{10}{5} = {r},
    cell{10}{6} = {r},
    cell{10}{7} = {r},
    cell{10}{8} = {r},
    cell{10}{9} = {r},
    cell{10}{10} = {r},
    cell{11}{2} = {r},
    cell{11}{3} = {r},
    cell{11}{4} = {r},
    cell{11}{5} = {r},
    cell{11}{6} = {r},
    cell{11}{7} = {r},
    cell{11}{8} = {r},
    cell{11}{9} = {r},
    cell{11}{10} = {r},
    cell{12}{2} = {r},
    cell{12}{3} = {r},
    cell{12}{4} = {r},
    cell{12}{5} = {r},
    cell{12}{6} = {r},
    cell{12}{7} = {r},
    cell{12}{8} = {r},
    cell{12}{9} = {r},
    cell{12}{10} = {r},
    cell{13}{2} = {r},
    cell{13}{3} = {r},
    cell{13}{4} = {r},
    cell{13}{5} = {r},
    cell{13}{6} = {r},
    cell{13}{7} = {r},
    cell{13}{8} = {r},
    cell{13}{9} = {r},
    cell{13}{10} = {r},
    cell{14}{2} = {c=18}{},
    cell{15}{2} = {c=6}{},
    cell{15}{8} = {c=12}{},
    cell{16}{2} = {c=6}{},
    cell{16}{8} = {c=6}{},
    cell{16}{14} = {c=6}{},
    cell{17}{2} = {c=3}{},
    cell{17}{5} = {c=3}{},
    cell{17}{8} = {c=3}{},
    cell{17}{11} = {c=3}{},
    cell{17}{14} = {c=3}{},
    cell{17}{17} = {c=3}{},
    cell{20}{2} = {r},
    cell{20}{3} = {r},
    cell{20}{4} = {r},
    cell{20}{5} = {r},
    cell{20}{6} = {r},
    cell{20}{7} = {r},
    cell{20}{8} = {r},
    cell{20}{9} = {r},
    cell{20}{10} = {r},
    cell{20}{11} = {r},
    cell{20}{12} = {r},
    cell{20}{13} = {r},
    cell{20}{14} = {r},
    cell{20}{15} = {r},
    cell{20}{16} = {r},
    cell{20}{17} = {r},
    cell{20}{18} = {r},
    cell{20}{19} = {r},
    cell{21}{2} = {r},
    cell{21}{3} = {r},
    cell{21}{4} = {r},
    cell{21}{5} = {r},
    cell{21}{6} = {r},
    cell{21}{7} = {r},
    cell{21}{8} = {r},
    cell{21}{9} = {r},
    cell{21}{10} = {r},
    cell{21}{11} = {r},
    cell{21}{12} = {r},
    cell{21}{13} = {r},
    cell{21}{14} = {r},
    cell{21}{15} = {r},
    cell{21}{16} = {r},
    cell{21}{17} = {r},
    cell{21}{18} = {r},
    cell{21}{19} = {r},
    cell{22}{2} = {r},
    cell{22}{3} = {r},
    cell{22}{4} = {r},
    cell{22}{5} = {r},
    cell{22}{6} = {r},
    cell{22}{7} = {r},
    cell{22}{8} = {r},
    cell{22}{9} = {r},
    cell{22}{10} = {r},
    cell{22}{11} = {r},
    cell{22}{12} = {r},
    cell{22}{13} = {r},
    cell{22}{14} = {r},
    cell{22}{15} = {r},
    cell{22}{16} = {r},
    cell{22}{17} = {r},
    cell{22}{18} = {r},
    cell{22}{19} = {r},
    cell{23}{2} = {r},
    cell{23}{3} = {r},
    cell{23}{4} = {r},
    cell{23}{5} = {r},
    cell{23}{6} = {r},
    cell{23}{7} = {r},
    cell{23}{8} = {r},
    cell{23}{9} = {r},
    cell{23}{10} = {r},
    cell{23}{11} = {r},
    cell{23}{12} = {r},
    cell{23}{13} = {r},
    cell{23}{14} = {r},
    cell{23}{15} = {r},
    cell{23}{16} = {r},
    cell{23}{17} = {r},
    cell{23}{18} = {r},
    cell{23}{19} = {r},
    cell{24}{2} = {r},
    cell{24}{3} = {r},
    cell{24}{4} = {r},
    cell{24}{5} = {r},
    cell{24}{6} = {r},
    cell{24}{7} = {r},
    cell{24}{8} = {r},
    cell{24}{9} = {r},
    cell{24}{10} = {r},
    cell{24}{11} = {r},
    cell{24}{12} = {r},
    cell{24}{13} = {r},
    cell{24}{14} = {r},
    cell{24}{15} = {r},
    cell{24}{16} = {r},
    cell{24}{17} = {r},
    cell{24}{18} = {r},
    cell{24}{19} = {r},
    cell{25}{2} = {r},
    cell{25}{3} = {r},
    cell{25}{4} = {r},
    cell{25}{5} = {r},
    cell{25}{6} = {r},
    cell{25}{7} = {r},
    cell{25}{8} = {r},
    cell{25}{9} = {r},
    cell{25}{10} = {r},
    cell{25}{11} = {r},
    cell{25}{12} = {r},
    cell{25}{13} = {r},
    cell{25}{14} = {r},
    cell{25}{15} = {r},
    cell{25}{16} = {r},
    cell{25}{17} = {r},
    cell{25}{18} = {r},
    cell{25}{19} = {r},
    cell{26}{2} = {r},
    cell{26}{3} = {r},
    cell{26}{4} = {r},
    cell{26}{5} = {r},
    cell{26}{6} = {r},
    cell{26}{7} = {r},
    cell{26}{8} = {r},
    cell{26}{9} = {r},
    cell{26}{10} = {r},
    cell{26}{11} = {r},
    cell{26}{12} = {r},
    cell{26}{13} = {r},
    cell{26}{14} = {r},
    cell{26}{15} = {r},
    cell{26}{16} = {r},
    cell{26}{17} = {r},
    cell{26}{18} = {r},
    cell{26}{19} = {r},
    cell{27}{2} = {c=18}{},
    cell{28}{2} = {c=6}{},
    cell{28}{8} = {c=12}{},
    cell{29}{2} = {c=6}{},
    cell{29}{8} = {c=6}{},
    cell{29}{14} = {c=6}{},
    cell{30}{2} = {c=3}{},
    cell{30}{5} = {c=3}{},
    cell{30}{8} = {c=3}{},
    cell{30}{11} = {c=3}{},
    cell{30}{14} = {c=3}{},
    cell{30}{17} = {c=3}{},
    cell{33}{2} = {r},
    cell{33}{3} = {r},
    cell{33}{4} = {r},
    cell{33}{5} = {r},
    cell{33}{6} = {r},
    cell{33}{7} = {r},
    cell{33}{8} = {r},
    cell{33}{9} = {r},
    cell{33}{10} = {r},
    cell{33}{11} = {r},
    cell{33}{12} = {r},
    cell{33}{13} = {r},
    cell{33}{14} = {r},
    cell{33}{15} = {r},
    cell{33}{16} = {r},
    cell{33}{17} = {r},
    cell{33}{18} = {r},
    cell{33}{19} = {r},
    cell{34}{2} = {r},
    cell{34}{3} = {r},
    cell{34}{4} = {r},
    cell{34}{5} = {r},
    cell{34}{6} = {r},
    cell{34}{7} = {r},
    cell{34}{8} = {r},
    cell{34}{9} = {r},
    cell{34}{10} = {r},
    cell{34}{11} = {r},
    cell{34}{12} = {r},
    cell{34}{13} = {r},
    cell{34}{14} = {r},
    cell{34}{15} = {r},
    cell{34}{16} = {r},
    cell{34}{17} = {r},
    cell{34}{18} = {r},
    cell{34}{19} = {r},
    cell{35}{2} = {r},
    cell{35}{3} = {r},
    cell{35}{4} = {r},
    cell{35}{5} = {r},
    cell{35}{6} = {r},
    cell{35}{7} = {r},
    cell{35}{8} = {r},
    cell{35}{9} = {r},
    cell{35}{10} = {r},
    cell{35}{11} = {r},
    cell{35}{12} = {r},
    cell{35}{13} = {r},
    cell{35}{14} = {r},
    cell{35}{15} = {r},
    cell{35}{16} = {r},
    cell{35}{17} = {r},
    cell{35}{18} = {r},
    cell{35}{19} = {r},
    cell{36}{2} = {r},
    cell{36}{3} = {r},
    cell{36}{4} = {r},
    cell{36}{5} = {r},
    cell{36}{6} = {r},
    cell{36}{7} = {r},
    cell{36}{8} = {r},
    cell{36}{9} = {r},
    cell{36}{10} = {r},
    cell{36}{11} = {r},
    cell{36}{12} = {r},
    cell{36}{13} = {r},
    cell{36}{14} = {r},
    cell{36}{15} = {r},
    cell{36}{16} = {r},
    cell{36}{17} = {r},
    cell{36}{18} = {r},
    cell{36}{19} = {r},
    cell{37}{2} = {r},
    cell{37}{3} = {r},
    cell{37}{4} = {r},
    cell{37}{5} = {r},
    cell{37}{6} = {r},
    cell{37}{7} = {r},
    cell{37}{8} = {r},
    cell{37}{9} = {r},
    cell{37}{10} = {r},
    cell{37}{11} = {r},
    cell{37}{12} = {r},
    cell{37}{13} = {r},
    cell{37}{14} = {r},
    cell{37}{15} = {r},
    cell{37}{16} = {r},
    cell{37}{17} = {r},
    cell{37}{18} = {r},
    cell{37}{19} = {r},
    cell{38}{2} = {r},
    cell{38}{3} = {r},
    cell{38}{4} = {r},
    cell{38}{5} = {r},
    cell{38}{6} = {r},
    cell{38}{7} = {r},
    cell{38}{8} = {r},
    cell{38}{9} = {r},
    cell{38}{10} = {r},
    cell{38}{11} = {r},
    cell{38}{12} = {r},
    cell{38}{13} = {r},
    cell{38}{14} = {r},
    cell{38}{15} = {r},
    cell{38}{16} = {r},
    cell{38}{17} = {r},
    cell{38}{18} = {r},
    cell{38}{19} = {r},
    cell{39}{2} = {r},
    cell{39}{3} = {r},
    cell{39}{4} = {r},
    cell{39}{5} = {r},
    cell{39}{6} = {r},
    cell{39}{7} = {r},
    cell{39}{8} = {r},
    cell{39}{9} = {r},
    cell{39}{10} = {r},
    cell{39}{11} = {r},
    cell{39}{12} = {r},
    cell{39}{13} = {r},
    cell{39}{14} = {r},
    cell{39}{15} = {r},
    cell{39}{16} = {r},
    cell{39}{17} = {r},
    cell{39}{18} = {r},
    cell{39}{19} = {r},
    vline{2} = {-}{},
    hline{1,7} = {1-10}{},
    hline{14,20,27,33,40} = {-}{},
  }
  \textbf{selection} & \textbf{GOM}          &               &               &               &               &               &               &               &               &               &               &               &               &               &               &               &               &               \\
  \textbf{approach}  & \textbf{GOMEA}        &               &               &               &               &               &               &               &               &               &               &               &               &               &               &               &               &               \\
  \textbf{cx}        & \textbf{LL-LT}        &               &               &               &               &               &               &               &               &               &               &               &               &               &               &               &               &               \\
  \textbf{(a)sync}   & \textbf{a/e}          &               &               & \textbf{a/i}  &               &               & \textbf{s}    &               &               &               &               &               &               &               &               &               &               &               \\
  \textbf{quantile}  & \textbf{0.25}         & \textbf{0.50} & \textbf{0.75} & \textbf{0.25} & \textbf{0.50} & \textbf{0.75} & \textbf{0.25} & \textbf{0.50} & \textbf{0.75} &               &               &               &               &               &               &               &               &               \\
  \textbf{timing}    &                       &               &               &               &               &               &               &               &               &               &               &               &               &               &               &               &               &               \\
  100:1              & 36                    & 44            & 53            & 36            & 44            & 53            & 36            & 44            & 53            &               &               &               &               &               &               &               &               &               \\
  10:1               & 36                    & 44            & 53            & 36            & 44            & 53            & 36            & 44            & 53            &               &               &               &               &               &               &               &               &               \\
  2:1                & 36                    & 44            & 53            & 36            & 44            & 53            & 36            & 44            & 53            &               &               &               &               &               &               &               &               &               \\
  1:1                & 36                    & 44            & 53            & 36            & 44            & 53            & 36            & 44            & 53            &               &               &               &               &               &               &               &               &               \\
  1:2                & 36                    & 44            & 56            & 36            & 44            & 53            & 36            & 44            & 53            &               &               &               &               &               &               &               &               &               \\
  1:10               & 36                    & 44            & 56            & 36            & 44            & 53            & 36            & 44            & 53            &               &               &               &               &               &               &               &               &               \\
  1:100              & 40                    & 52            & 64            & 36            & 44            & 53            & 36            & 44            & 53            &               &               &               &               &               &               &               &               &               \\
  \textbf{selection} & \textbf{steady-state} &               &               &               &               &               &               &               &               &               &               &               &               &               &               &               &               &               \\
  \textbf{approach}  & \textbf{ECGA}         &               &               &               &               &               & \textbf{GA}   &               &               &               &               &               &               &               &               &               &               &               \\
  \textbf{cx}        & \textbf{LL-MPM}       &               &               &               &               &               & \textbf{TPX}  &               &               &               &               &               & \textbf{SFX}  &               &               &               &               &               \\
  \textbf{(a)sync}   & \textbf{a}            &               &               & \textbf{s}    &               &               & \textbf{a}    &               &               & \textbf{s}    &               &               & \textbf{a}    &               &               & \textbf{s}    &               &               \\
  \textbf{quantile}  & \textbf{0.25}         & \textbf{0.50} & \textbf{0.75} & \textbf{0.25} & \textbf{0.50} & \textbf{0.75} & \textbf{0.25} & \textbf{0.50} & \textbf{0.75} & \textbf{0.25} & \textbf{0.50} & \textbf{0.75} & \textbf{0.25} & \textbf{0.50} & \textbf{0.75} & \textbf{0.25} & \textbf{0.50} & \textbf{0.75} \\
  \textbf{timing}    &                       &               &               &               &               &               &               &               &               &               &               &               &               &               &               &               &               &               \\
  100:1              & 1787                  & 1944          & 2100          & 4024          & 4216          & 4655          & 152           & 192           & 241           & 216           & 262           & 304           & 80            & 102           & 128           & 108           & 132           & 158           \\
  10:1               & 1909                  & 2038          & 2136          & 4024          & 4216          & 4655          & 160           & 196           & 238           & 216           & 262           & 304           & 87            & 102           & 128           & 108           & 132           & 158           \\
  2:1                & 2014                  & 2166          & 2365          & 4024          & 4216          & 4655          & 160           & 208           & 246           & 216           & 262           & 304           & 76            & 104           & 116           & 108           & 132           & 158           \\
  1:1                & 3894                  & 4110          & 4612          & 3894          & 4110          & 4612          & 205           & 264           & 329           & 205           & 264           & 329           & 96            & 130           & 161           & 96            & 130           & 161           \\
  1:2                & 1975                  & 2110          & 2313          & 3955          & 4118          & 4632          & 206           & 256           & 328           & 192           & 244           & 304           & 96            & 120           & 148           & 96            & 122           & 148           \\
  1:10               & 2226                  & 2394          & 2593          & 3955          & 4118          & 4632          & 243           & 286           & 341           & 192           & 244           & 304           & 104           & 132           & 152           & 96            & 122           & 148           \\
  1:100              & 2396                  & 2562          & 2756          & 3955          & 4118          & 4632          & 239           & 288           & 356           & 192           & 244           & 304           & 100           & 128           & 156           & 96            & 122           & 148           \\
  \textbf{selection} & \textbf{generational} &               &               &               &               &               &               &               &               &               &               &               &               &               &               &               &               &               \\
  \textbf{approach}  & \textbf{ECGA}         &               &               &               &               &               & \textbf{GA}   &               &               &               &               &               &               &               &               &               &               &               \\
  \textbf{cx}        & \textbf{LL-MPM}       &               &               &               &               &               & \textbf{TPX}  &               &               &               &               &               & \textbf{SFX}  &               &               &               &               &               \\
  \textbf{(a)sync}   & \textbf{a}            &               &               & \textbf{s}    &               &               & \textbf{a}    &               &               & \textbf{s}    &               &               & \textbf{a}    &               &               & \textbf{s}    &               &               \\
  \textbf{quantile}  & \textbf{0.25}         & \textbf{0.50} & \textbf{0.75} & \textbf{0.25} & \textbf{0.50} & \textbf{0.75} & \textbf{0.25} & \textbf{0.50} & \textbf{0.75} & \textbf{0.25} & \textbf{0.50} & \textbf{0.75} & \textbf{0.25} & \textbf{0.50} & \textbf{0.75} & \textbf{0.25} & \textbf{0.50} & \textbf{0.75} \\
  \textbf{timing}    &                       &               &               &               &               &               &               &               &               &               &               &               &               &               &               &               &               &               \\
  100:1              & 1753                  & 1906          & 2085          & 3759          & 4046          & 4380          & 316           & 434           & 537           & 403           & 546           & 684           & 128           & 176           & 204           & 152           & 190           & 224           \\
  10:1               & 1824                  & 2022          & 2173          & 3759          & 4046          & 4380          & 319           & 450           & 558           & 403           & 546           & 684           & 147           & 174           & 212           & 152           & 190           & 224           \\
  2:1                & 1915                  & 2042          & 2180          & 3759          & 4046          & 4380          & 422           & 512           & 661           & 403           & 546           & 684           & 128           & 164           & 228           & 152           & 190           & 224           \\
  1:1                & 3712                  & 4052          & 4269          & 3712          & 4052          & 4269          & 435           & 496           & 646           & 435           & 496           & 646           & 143           & 182           & 248           & 143           & 182           & 248           \\
  1:2                & 1920                  & 2060          & 2222          & 3823          & 4050          & 4492          & 320           & 482           & 598           & 384           & 508           & 637           & 128           & 168           & 218           & 152           & 188           & 220           \\
  1:10               & 2149                  & 2316          & 2528          & 3823          & 4050          & 4492          & 451           & 584           & 772           & 384           & 508           & 637           & 155           & 198           & 240           & 152           & 188           & 220           \\
  1:100              & 2304                  & 2534          & 2693          & 3823          & 4050          & 4492          & 507           & 634           & 863           & 384           & 508           & 637           & 180           & 234           & 268           & 152           & 188           & 220           
  \end{tblr}
  \vspace*{1mm}
  \captionof{table}{Quantiles of minimally required population sizes for DT}\label{tab:table-dt-ext}

\small
\begin{tblr}{
    cell{1}{2} = {c=9}{},
    cell{2}{2} = {c=9}{},
    cell{3}{2} = {c=9}{},
    cell{4}{2} = {c=3}{},
    cell{4}{5} = {c=3}{},
    cell{4}{8} = {c=3}{},
    cell{7}{2} = {r},
    cell{7}{3} = {r},
    cell{7}{4} = {r},
    cell{7}{5} = {r},
    cell{7}{6} = {r},
    cell{7}{7} = {r},
    cell{7}{8} = {r},
    cell{7}{9} = {r},
    cell{7}{10} = {r},
    cell{8}{2} = {r},
    cell{8}{3} = {r},
    cell{8}{4} = {r},
    cell{8}{5} = {r},
    cell{8}{6} = {r},
    cell{8}{7} = {r},
    cell{8}{8} = {r},
    cell{8}{9} = {r},
    cell{8}{10} = {r},
    cell{9}{2} = {r},
    cell{9}{3} = {r},
    cell{9}{4} = {r},
    cell{9}{5} = {r},
    cell{9}{6} = {r},
    cell{9}{7} = {r},
    cell{9}{8} = {r},
    cell{9}{9} = {r},
    cell{9}{10} = {r},
    cell{10}{2} = {r},
    cell{10}{3} = {r},
    cell{10}{4} = {r},
    cell{10}{5} = {r},
    cell{10}{6} = {r},
    cell{10}{7} = {r},
    cell{10}{8} = {r},
    cell{10}{9} = {r},
    cell{10}{10} = {r},
    cell{11}{2} = {r},
    cell{11}{3} = {r},
    cell{11}{4} = {r},
    cell{11}{5} = {r},
    cell{11}{6} = {r},
    cell{11}{7} = {r},
    cell{11}{8} = {r},
    cell{11}{9} = {r},
    cell{11}{10} = {r},
    cell{12}{2} = {r},
    cell{12}{3} = {r},
    cell{12}{4} = {r},
    cell{12}{5} = {r},
    cell{12}{6} = {r},
    cell{12}{7} = {r},
    cell{12}{8} = {r},
    cell{12}{9} = {r},
    cell{12}{10} = {r},
    cell{13}{2} = {r},
    cell{13}{3} = {r},
    cell{13}{4} = {r},
    cell{13}{5} = {r},
    cell{13}{6} = {r},
    cell{13}{7} = {r},
    cell{13}{8} = {r},
    cell{13}{9} = {r},
    cell{13}{10} = {r},
    cell{14}{2} = {c=24}{},
    cell{15}{2} = {c=6}{},
    cell{15}{8} = {c=18}{},
    cell{16}{2} = {c=6}{},
    cell{16}{8} = {c=6}{},
    cell{16}{14} = {c=6}{},
    cell{16}{20} = {c=6}{},
    cell{17}{2} = {c=3}{},
    cell{17}{5} = {c=3}{},
    cell{17}{8} = {c=3}{},
    cell{17}{11} = {c=3}{},
    cell{17}{14} = {c=3}{},
    cell{17}{17} = {c=3}{},
    cell{17}{20} = {c=3}{},
    cell{17}{23} = {c=3}{},
    cell{20}{2} = {r},
    cell{20}{3} = {r},
    cell{20}{4} = {r},
    cell{20}{5} = {r},
    cell{20}{6} = {r},
    cell{20}{7} = {r},
    cell{20}{8} = {r},
    cell{20}{9} = {r},
    cell{20}{10} = {r},
    cell{20}{11} = {r},
    cell{20}{12} = {r},
    cell{20}{13} = {r},
    cell{20}{14} = {r},
    cell{20}{15} = {r},
    cell{20}{16} = {r},
    cell{20}{17} = {r},
    cell{20}{18} = {r},
    cell{20}{19} = {r},
    cell{20}{20} = {r},
    cell{20}{21} = {r},
    cell{20}{22} = {r},
    cell{20}{23} = {r},
    cell{20}{24} = {r},
    cell{20}{25} = {r},
    cell{21}{2} = {r},
    cell{21}{3} = {r},
    cell{21}{4} = {r},
    cell{21}{5} = {r},
    cell{21}{6} = {r},
    cell{21}{7} = {r},
    cell{21}{8} = {r},
    cell{21}{9} = {r},
    cell{21}{10} = {r},
    cell{21}{11} = {r},
    cell{21}{12} = {r},
    cell{21}{13} = {r},
    cell{21}{14} = {r},
    cell{21}{15} = {r},
    cell{21}{16} = {r},
    cell{21}{17} = {r},
    cell{21}{18} = {r},
    cell{21}{19} = {r},
    cell{21}{20} = {r},
    cell{21}{21} = {r},
    cell{21}{22} = {r},
    cell{21}{23} = {r},
    cell{21}{24} = {r},
    cell{21}{25} = {r},
    cell{22}{2} = {r},
    cell{22}{3} = {r},
    cell{22}{4} = {r},
    cell{22}{5} = {r},
    cell{22}{6} = {r},
    cell{22}{7} = {r},
    cell{22}{8} = {r},
    cell{22}{9} = {r},
    cell{22}{10} = {r},
    cell{22}{11} = {r},
    cell{22}{12} = {r},
    cell{22}{13} = {r},
    cell{22}{14} = {r},
    cell{22}{15} = {r},
    cell{22}{16} = {r},
    cell{22}{17} = {r},
    cell{22}{18} = {r},
    cell{22}{19} = {r},
    cell{22}{20} = {r},
    cell{22}{21} = {r},
    cell{22}{22} = {r},
    cell{22}{23} = {r},
    cell{22}{24} = {r},
    cell{22}{25} = {r},
    cell{23}{2} = {r},
    cell{23}{3} = {r},
    cell{23}{4} = {r},
    cell{23}{5} = {r},
    cell{23}{6} = {r},
    cell{23}{7} = {r},
    cell{23}{8} = {r},
    cell{23}{9} = {r},
    cell{23}{10} = {r},
    cell{23}{14} = {r},
    cell{23}{15} = {r},
    cell{23}{16} = {r},
    cell{23}{17} = {r},
    cell{23}{18} = {r},
    cell{23}{19} = {r},
    cell{23}{20} = {r},
    cell{23}{21} = {r},
    cell{23}{22} = {r},
    cell{23}{23} = {r},
    cell{23}{24} = {r},
    cell{23}{25} = {r},
    cell{24}{2} = {r},
    cell{24}{3} = {r},
    cell{24}{4} = {r},
    cell{24}{5} = {r},
    cell{24}{6} = {r},
    cell{24}{7} = {r},
    cell{24}{8} = {r},
    cell{24}{9} = {r},
    cell{24}{10} = {r},
    cell{24}{14} = {r},
    cell{24}{15} = {r},
    cell{24}{16} = {r},
    cell{24}{17} = {r},
    cell{24}{18} = {r},
    cell{24}{19} = {r},
    cell{24}{20} = {r},
    cell{24}{21} = {r},
    cell{24}{22} = {r},
    cell{24}{23} = {r},
    cell{24}{24} = {r},
    cell{25}{2} = {r},
    cell{25}{3} = {r},
    cell{25}{4} = {r},
    cell{25}{5} = {r},
    cell{25}{6} = {r},
    cell{25}{7} = {r},
    cell{25}{8} = {r},
    cell{25}{9} = {r},
    cell{25}{10} = {r},
    cell{25}{14} = {r},
    cell{25}{15} = {r},
    cell{25}{16} = {r},
    cell{25}{17} = {r},
    cell{25}{18} = {r},
    cell{25}{19} = {r},
    cell{25}{20} = {r},
    cell{25}{21} = {r},
    cell{25}{22} = {r},
    cell{25}{23} = {r},
    cell{25}{24} = {r},
    cell{26}{2} = {r},
    cell{26}{3} = {r},
    cell{26}{4} = {r},
    cell{26}{5} = {r},
    cell{26}{6} = {r},
    cell{26}{7} = {r},
    cell{26}{8} = {r},
    cell{26}{9} = {r},
    cell{26}{10} = {r},
    cell{26}{14} = {r},
    cell{26}{15} = {r},
    cell{26}{16} = {r},
    cell{26}{17} = {r},
    cell{26}{18} = {r},
    cell{26}{19} = {r},
    cell{26}{20} = {r},
    cell{26}{21} = {r},
    cell{26}{22} = {r},
    cell{26}{23} = {r},
    cell{27}{2} = {c=24}{},
    cell{28}{2} = {c=6}{},
    cell{28}{8} = {c=18}{},
    cell{29}{2} = {c=6}{},
    cell{29}{8} = {c=6}{},
    cell{29}{14} = {c=6}{},
    cell{29}{20} = {c=6}{},
    cell{30}{2} = {c=3}{},
    cell{30}{5} = {c=3}{},
    cell{30}{8} = {c=3}{},
    cell{30}{11} = {c=3}{},
    cell{30}{14} = {c=3}{},
    cell{30}{17} = {c=3}{},
    cell{30}{20} = {c=3}{},
    cell{30}{23} = {c=3}{},
    cell{33}{2} = {r},
    cell{33}{3} = {r},
    cell{33}{4} = {r},
    cell{33}{5} = {r},
    cell{33}{6} = {r},
    cell{33}{7} = {r},
    cell{33}{8} = {r},
    cell{33}{9} = {r},
    cell{33}{10} = {r},
    cell{33}{11} = {r},
    cell{33}{12} = {r},
    cell{33}{13} = {r},
    cell{33}{14} = {r},
    cell{33}{15} = {r},
    cell{33}{16} = {r},
    cell{33}{17} = {r},
    cell{33}{18} = {r},
    cell{33}{19} = {r},
    cell{33}{20} = {r},
    cell{33}{21} = {r},
    cell{33}{22} = {r},
    cell{33}{23} = {r},
    cell{33}{24} = {r},
    cell{33}{25} = {r},
    cell{34}{2} = {r},
    cell{34}{3} = {r},
    cell{34}{4} = {r},
    cell{34}{5} = {r},
    cell{34}{6} = {r},
    cell{34}{7} = {r},
    cell{34}{8} = {r},
    cell{34}{9} = {r},
    cell{34}{10} = {r},
    cell{34}{11} = {r},
    cell{34}{12} = {r},
    cell{34}{13} = {r},
    cell{34}{14} = {r},
    cell{34}{15} = {r},
    cell{34}{16} = {r},
    cell{34}{17} = {r},
    cell{34}{18} = {r},
    cell{34}{19} = {r},
    cell{34}{20} = {r},
    cell{34}{21} = {r},
    cell{34}{22} = {r},
    cell{34}{23} = {r},
    cell{34}{24} = {r},
    cell{34}{25} = {r},
    cell{35}{2} = {r},
    cell{35}{3} = {r},
    cell{35}{4} = {r},
    cell{35}{5} = {r},
    cell{35}{6} = {r},
    cell{35}{7} = {r},
    cell{35}{8} = {r},
    cell{35}{9} = {r},
    cell{35}{10} = {r},
    cell{35}{11} = {r},
    cell{35}{12} = {r},
    cell{35}{13} = {r},
    cell{35}{14} = {r},
    cell{35}{15} = {r},
    cell{35}{16} = {r},
    cell{35}{17} = {r},
    cell{35}{18} = {r},
    cell{35}{19} = {r},
    cell{35}{20} = {r},
    cell{35}{21} = {r},
    cell{35}{22} = {r},
    cell{35}{23} = {r},
    cell{35}{24} = {r},
    cell{35}{25} = {r},
    cell{36}{2} = {r},
    cell{36}{3} = {r},
    cell{36}{4} = {r},
    cell{36}{5} = {r},
    cell{36}{6} = {r},
    cell{36}{7} = {r},
    cell{36}{8} = {r},
    cell{36}{9} = {r},
    cell{36}{10} = {r},
    cell{36}{11} = {r},
    cell{36}{12} = {r},
    cell{36}{13} = {r},
    cell{36}{14} = {r},
    cell{36}{15} = {r},
    cell{36}{16} = {r},
    cell{36}{17} = {r},
    cell{36}{18} = {r},
    cell{36}{19} = {r},
    cell{36}{20} = {r},
    cell{36}{21} = {r},
    cell{36}{22} = {r},
    cell{36}{23} = {r},
    cell{36}{24} = {r},
    cell{36}{25} = {r},
    cell{37}{2} = {r},
    cell{37}{3} = {r},
    cell{37}{4} = {r},
    cell{37}{5} = {r},
    cell{37}{6} = {r},
    cell{37}{7} = {r},
    cell{37}{8} = {r},
    cell{37}{9} = {r},
    cell{37}{10} = {r},
    cell{37}{11} = {r},
    cell{37}{12} = {r},
    cell{37}{13} = {r},
    cell{37}{14} = {r},
    cell{37}{15} = {r},
    cell{37}{16} = {r},
    cell{37}{17} = {r},
    cell{37}{18} = {r},
    cell{37}{19} = {r},
    cell{37}{20} = {r},
    cell{37}{21} = {r},
    cell{37}{22} = {r},
    cell{37}{23} = {r},
    cell{37}{24} = {r},
    cell{37}{25} = {r},
    cell{38}{2} = {r},
    cell{38}{3} = {r},
    cell{38}{4} = {r},
    cell{38}{5} = {r},
    cell{38}{6} = {r},
    cell{38}{7} = {r},
    cell{38}{8} = {r},
    cell{38}{9} = {r},
    cell{38}{10} = {r},
    cell{38}{11} = {r},
    cell{38}{12} = {r},
    cell{38}{13} = {r},
    cell{38}{14} = {r},
    cell{38}{15} = {r},
    cell{38}{16} = {r},
    cell{38}{17} = {r},
    cell{38}{18} = {r},
    cell{38}{19} = {r},
    cell{38}{20} = {r},
    cell{38}{21} = {r},
    cell{38}{22} = {r},
    cell{38}{23} = {r},
    cell{38}{24} = {r},
    cell{38}{25} = {r},
    cell{39}{2} = {r},
    cell{39}{3} = {r},
    cell{39}{4} = {r},
    cell{39}{5} = {r},
    cell{39}{6} = {r},
    cell{39}{7} = {r},
    cell{39}{8} = {r},
    cell{39}{9} = {r},
    cell{39}{10} = {r},
    cell{39}{11} = {r},
    cell{39}{12} = {r},
    cell{39}{13} = {r},
    cell{39}{14} = {r},
    cell{39}{15} = {r},
    cell{39}{16} = {r},
    cell{39}{17} = {r},
    cell{39}{18} = {r},
    cell{39}{19} = {r},
    cell{39}{20} = {r},
    cell{39}{21} = {r},
    cell{39}{22} = {r},
    cell{39}{23} = {r},
    cell{39}{24} = {r},
    cell{39}{25} = {r},
    vline{2} = {1-39}{},
    hline{1,7} = {1-10}{},
    hline{14,20,27,33,40} = {-}{},
}
\textbf{selection} & \textbf{GOM}            &               &               &               &               &               &               &               &               &               &               &               &               &               &               &               &               &               &               &               &               &               &               &               \\
\textbf{approach}  & \textbf{GOMEA}        &               &               &               &               &               &               &               &               &               &               &               &               &               &               &               &               &               &               &               &               &               &               &               \\
\textbf{cx}        & \textbf{LL-LT}            &               &               &               &               &               &               &               &               &               &               &               &               &               &               &               &               &               &               &               &               &               &               &               \\
\textbf{(a)sync}   & \textbf{a/e}          &               &               & \textbf{a/i}  &               &               & \textbf{s}    &               &               &               &               &               &               &               &               &               &               &               &               &               &               &               &               &               \\
\textbf{quantile}  & \textbf{0.25}         & \textbf{0.50} & \textbf{0.75} & \textbf{0.25} & \textbf{0.50} & \textbf{0.75} & \textbf{0.25} & \textbf{0.50} & \textbf{0.75} &               &               &               &               &               &               &               &               &               &               &               &               &               &               &               \\
\textbf{timing}    &                       &               &               &               &               &               &               &               &               &               &               &               &               &               &               &               &               &               &               &               &               &               &               &               \\
100:1              & 32                    & 40            & 56            & 32            & 40            & 52            & 32            & 44            & 65            &               &               &               &               &               &               &               &               &               &               &               &               &               &               &               \\
10:1               & 32                    & 40            & 60            & 32            & 36            & 52            & 32            & 48            & 60            &               &               &               &               &               &               &               &               &               &               &               &               &               &               &               \\
2:1                & 32                    & 40            & 53            & 32            & 40            & 56            & 32            & 44            & 56            &               &               &               &               &               &               &               &               &               &               &               &               &               &               &               \\
1:1                & 32                    & 48            & 64            & 32            & 48            & 64            & 32            & 44            & 64            &               &               &               &               &               &               &               &               &               &               &               &               &               &               &               \\
1:2                & 32                    & 42            & 60            & 32            & 40            & 56            & 36            & 48            & 68            &               &               &               &               &               &               &               &               &               &               &               &               &               &               &               \\
1:10               & 32                    & 40            & 56            & 32            & 40            & 56            & 32            & 48            & 57            &               &               &               &               &               &               &               &               &               &               &               &               &               &               &               \\
1:100              & 32                    & 40            & 56            & 32            & 40            & 56            & 32            & 48            & 64            &               &               &               &               &               &               &               &               &               &               &               &               &               &               &               \\
\textbf{selection} & \textbf{steady-state} &               &               &               &               &               &               &               &               &               &               &               &               &               &               &               &               &               &               &               &               &               &               &               \\
\textbf{approach}  & \textbf{ECGA}         &               &               &               &               &               & \textbf{GA}   &               &               &               &               &               &               &               &               &               &               &               &               &               &               &               &               &               \\
\textbf{cx}        & \textbf{LL-MPM}            &               &               &               &               &               & \textbf{UX}   &               &               &               &               &               & \textbf{TPX}  &               &               &               &               &               & \textbf{SFX}  &               &               &               &               &               \\
\textbf{(a)sync}   & \textbf{s}            &               &               & \textbf{a}    &               &               & \textbf{s}    &               &               & \textbf{a}    &               &               & \textbf{s}    &               &               & \textbf{a}    &               &               & \textbf{s}    &               &               & \textbf{a}    &               &               \\
\textbf{quantile}  & \textbf{0.25}         & \textbf{0.50} & \textbf{0.75} & \textbf{0.25} & \textbf{0.50} & \textbf{0.75} & \textbf{0.25} & \textbf{0.50} & \textbf{0.75} & \textbf{0.25} & \textbf{0.50} & \textbf{0.75} & \textbf{0.25} & \textbf{0.50} & \textbf{0.75} & \textbf{0.25} & \textbf{0.50} & \textbf{0.75} & \textbf{0.25} & \textbf{0.50} & \textbf{0.75} & \textbf{0.25} & \textbf{0.50} & \textbf{0.75} \\
\textbf{timing}    &                       &               &               &               &               &               &               &               &               &               &               &               &               &               &               &               &               &               &               &               &               &               &               &               \\
100:1              & 1661                  & 4004          & 6563          & 236           & 454           & 732           & 4096          & 16260         & 31800         & 1020          & 7034          & 19880         & 124           & 206           & 337           & 72            & 120           & 164           & 3252          & 11516         & 18992         & 183           & 968           & 4305          \\
10:1               & 1661                  & 4004          & 6563          & 249           & 506           & 897           & 4096          & 16260         & 31800         & 499           & 8156          & 21943         & 124           & 206           & 337           & 64            & 114           & 161           & 3252          & 11516         & 18992         & 406           & 3602          & 7900          \\
2:1                & 1661                  & 4004          & 6563          & 576           & 1426          & 3004          & 4096          & 16260         & 31800         & 1280          & 16380         & 49152         & 124           & 206           & 337           & 88            & 124           & 190           & 3252          & 11516         & 18992         & 1697          & 7160          & 14676         \\
1:1                & 1590                  & 2992          & 6667          & 1590          & 2992          & 6667          & 4096          & 17098         & 30462         &               &               &               & 115           & 178           & 260           & 115           & 178           & 260           & 4095          & 9540          & 17835         & 8184          & 23488         & 46400         \\
1:2                & 1747                  & 3760          & 6286          & 892           & 1784          & 2861          & 4096          & 15648         & 27424         &               &               &               & 123           & 182           & 264           & 112           & 194           & 277           & 4094          & 10848         & 17519         & 16270         & 49152         &               \\
1:10               & 1747                  & 3760          & 6286          & 1628          & 2924          & 4937          & 4096          & 15648         & 27424         &               &               &               & 123           & 182           & 264           & 159           & 228           & 318           & 4094          & 10848         & 17519         & 24544         & 57344         &               \\
1:100              & 1747                  & 3760          & 6286          & 1951          & 3722          & 6256          & 4096          & 15648         & 27424         &               &               &               & 123           & 182           & 264           & 128           & 218           & 305           & 4094          & 10848         & 17519         & 31392         &               &               \\
\textbf{selection} & \textbf{generational} &               &               &               &               &               &               &               &               &               &               &               &               &               &               &               &               &               &               &               &               &               &               &               \\
\textbf{approach}  & \textbf{ECGA}         &               &               &               &               &               & \textbf{GA}   &               &               &               &               &               &               &               &               &               &               &               &               &               &               &               &               &               \\
\textbf{cx}        & \textbf{LL-MPM}            &               &               &               &               &               & \textbf{UX}   &               &               &               &               &               & \textbf{TPX}  &               &               &               &               &               & \textbf{SFX}  &               &               &               &               &               \\
\textbf{(a)sync}   & \textbf{s}            &               &               & \textbf{a}    &               &               & \textbf{s}    &               &               & \textbf{a}    &               &               & \textbf{s}    &               &               & \textbf{a}    &               &               & \textbf{s}    &               &               & \textbf{a}    &               &               \\
\textbf{quantile}  & \textbf{0.25}         & \textbf{0.50} & \textbf{0.75} & \textbf{0.25} & \textbf{0.50} & \textbf{0.75} & \textbf{0.25} & \textbf{0.50} & \textbf{0.75} & \textbf{0.25} & \textbf{0.50} & \textbf{0.75} & \textbf{0.25} & \textbf{0.50} & \textbf{0.75} & \textbf{0.25} & \textbf{0.50} & \textbf{0.75} & \textbf{0.25} & \textbf{0.50} & \textbf{0.75} & \textbf{0.25} & \textbf{0.50} & \textbf{0.75} \\
\textbf{timing} &                       &               &               &               &               &               &               &               &               &               &               &               &               &               &               &               &               &               &               &               &               &               &               &               \\
100:1              & 1536                  & 3704          & 6280          & 202           & 382           & 576           & 1904          & 3682          & 6182          & 512           & 1968          & 4051          & 243           & 408           & 603           & 182           & 268           & 393           & 1531          & 2990          & 5997          & 743           & 1578          & 2689          \\
10:1               & 1536                  & 3704          & 6280          & 256           & 512           & 902           & 1904          & 3682          & 6182          & 981           & 1982          & 3958          & 243           & 408           & 603           & 137           & 254           & 432           & 1531          & 2990          & 5997          & 955           & 1752          & 3695          \\
2:1                & 1536                  & 3704          & 6280          & 635           & 1214          & 2288          & 1904          & 3682          & 6182          & 1024          & 3468          & 6749          & 243           & 408           & 603           & 236           & 396           & 573           & 1531          & 2990          & 5997          & 1745          & 3634          & 5827          \\
1:1                & 1705                  & 4012          & 6232          & 1705          & 4012          & 6232          & 2615          & 3962          & 7007          & 2909          & 4076          & 7721          & 252           & 386           & 600           & 252           & 386           & 606           & 1024          & 3058          & 5944          & 1024          & 3442          & 6684          \\
1:2                & 1390                  & 2848          & 6904          & 933           & 1756          & 2576          & 1837          & 3630          & 7217          & 2015          & 4066          & 7544          & 254           & 444           & 689           & 252           & 444           & 614           & 1791          & 3310          & 6813          & 1498          & 2872          & 5738          \\
1:10               & 1390                  & 2848          & 6904          & 1486          & 2802          & 5138          & 1837          & 3630          & 7217          & 3037          & 6260          & 11703         & 254           & 444           & 689           & 332           & 516           & 911           & 1791          & 3310          & 6813          & 2048          & 5046          & 8492          \\
1:100              & 1390                  & 2848          & 6904          & 1764          & 3654          & 6338          & 1837          & 3630          & 7217          & 3748          & 8148          & 15482         & 254           & 444           & 689           & 256           & 512           & 860           & 1791          & 3310          & 6813          & 1887          & 4096          & 8480          
\end{tblr}
\vspace*{1mm}
\captionof{table}{Quantiles of minimally required population sizes for ANKL}
\label{tab:table-ankl-ext}
\end{landscape}
\restoregeometry

\begin{table}
    \caption{Quantiles of minimally required time and corresponding population sizes for NASBench 301}
    \label{tab:table-nasbench-ext}
    \begin{tblr}{
  cell{1}{5} = {c=3}{},
  cell{1}{8} = {c=3}{},
  cell{4}{1} = {r=3}{},
  cell{4}{2} = {r=3}{},
  cell{4}{3} = {r=3}{},
  cell{4}{5} = {r},
  cell{4}{6} = {r},
  cell{4}{7} = {r},
  cell{4}{8} = {r},
  cell{4}{9} = {r},
  cell{4}{10} = {r},
  cell{5}{5} = {r},
  cell{5}{6} = {r},
  cell{5}{7} = {r},
  cell{5}{8} = {r},
  cell{5}{9} = {r},
  cell{5}{10} = {r},
  cell{6}{5} = {r},
  cell{6}{6} = {r},
  cell{6}{7} = {r},
  cell{6}{8} = {r},
  cell{6}{9} = {r},
  cell{6}{10} = {r},
  cell{7}{1} = {r=6}{},
  cell{7}{2} = {r=2}{},
  cell{7}{3} = {r=2}{},
  cell{9}{2} = {r=4}{},
  cell{9}{3} = {r=2}{},
  cell{9}{5} = {r},
  cell{9}{6} = {r},
  cell{9}{7} = {r},
  cell{9}{8} = {r},
  cell{9}{9} = {r},
  cell{9}{10} = {r},
  cell{10}{5} = {r},
  cell{10}{6} = {r},
  cell{10}{7} = {r},
  cell{10}{8} = {r},
  cell{10}{9} = {r},
  cell{10}{10} = {r},
  cell{11}{3} = {r=2}{},
  cell{11}{5} = {r},
  cell{11}{6} = {r},
  cell{11}{7} = {r},
  cell{11}{8} = {r},
  cell{11}{9} = {r},
  cell{11}{10} = {r},
  cell{12}{5} = {r},
  cell{12}{6} = {r},
  cell{12}{7} = {r},
  cell{12}{8} = {r},
  cell{12}{9} = {r},
  cell{12}{10} = {r},
  cell{13}{1} = {r=6}{},
  cell{13}{2} = {r=2}{},
  cell{13}{3} = {r=2}{},
  cell{13}{5} = {r},
  cell{13}{6} = {r},
  cell{13}{7} = {r},
  cell{13}{8} = {r},
  cell{13}{9} = {r},
  cell{13}{10} = {r},
  cell{15}{2} = {r=4}{},
  cell{15}{3} = {r=2}{},
  cell{15}{5} = {r},
  cell{15}{6} = {r},
  cell{15}{7} = {r},
  cell{15}{8} = {r},
  cell{15}{9} = {r},
  cell{15}{10} = {r},
  cell{16}{5} = {r},
  cell{16}{6} = {r},
  cell{16}{7} = {r},
  cell{16}{8} = {r},
  cell{16}{9} = {r},
  cell{16}{10} = {r},
  cell{17}{3} = {r=2}{},
  cell{17}{5} = {r},
  cell{17}{6} = {r},
  cell{17}{7} = {r},
  cell{17}{8} = {r},
  cell{17}{9} = {r},
  cell{17}{10} = {r},
  cell{18}{5} = {r},
  cell{18}{6} = {r},
  cell{18}{7} = {r},
  cell{18}{8} = {r},
  cell{18}{9} = {r},
  cell{18}{10} = {r},
  vline{5,8} = {1-18}{},
  hline{1,4,19} = {-}{},
}
                      &                   &                 & ~                & \textbf{simulation time (s) × $\mathbf{10^6}$} &               &               & \textbf{population\_size} &               &               \\
                      &                   &                 & ~                & \textbf{0.25}                        & \textbf{0.50} & \textbf{0.75} & \textbf{0.25}             & \textbf{0.50} & \textbf{0.75} \\
\textbf{selection}    & \textbf{approach} & \textbf{cx}     & \textbf{(a)sync} &                                      &               &               &                           &               &               \\
\textbf{GOM}          & \textbf{GOMEA}    & \textbf{LL-LT}  & \textbf{a/e}     & 1.591601                             & 1.677091      & 1.786227      & 38.75                     & 49.5          & 64            \\
                      &                   &                 & \textbf{a/i}     & 1.215592                             & 1.361821      & 1.64783       & 46                        & 63.5          & 79.5          \\
                      &                   &                 & \textbf{s}       & 1.462275                             & 1.652285      & 1.788804      & 48.5                      & 58.5          & 62.5          \\
\textbf{steady-state} & \textbf{ECGA}     & \textbf{LL-MPM} & \textbf{a}       &                                      &               &               &                           &               &               \\
                      &                   &                 & \textbf{s}       &                                      &               &               &                           &               &               \\
                      & \textbf{GA}       & \textbf{UX}     & \textbf{a}       & 0.75999                              & 0.982413      & 1.340313      & 208.5                     & 252.5         & 372.75        \\
                      &                   &                 & \textbf{s}       & 1.125931                             & 1.400998      & 1.790128      & 256                       & 368.5         & 512           \\
                      &                   & \textbf{TPX}    & \textbf{a}       & 3.672774                             & 4.258886      & 7.41098       & 960.25                    & 1024          & 2048          \\
                      &                   &                 & \textbf{s}       & 3.761896                             & 4.492289      & 8.453775      & 767.75                    & 1110          & 2048          \\
\textbf{generational} & \textbf{ECGA}     & \textbf{LL-MPM} & \textbf{a}       & 2.515486                             & 7.734439      & 12.016        & 1024                      & 3072          & 4096          \\
                      &                   &                 & \textbf{s}       &                                      &               &               &                           &               &               \\
                      & \textbf{GA}       & \textbf{UX}     & \textbf{a}       & 1.20572                              & 1.517906      & 1.959241      & 514.25                    & 816           & 1024          \\
                      &                   &                 & \textbf{s}       & 0.746899                             & 0.980716      & 2.065738      & 274.25                    & 480.5         & 1052.5        \\
                      &                   & \textbf{TPX}    & \textbf{a}       & 2.583693                             & 4.434104      & 6.377228      & 1024                      & 2048          & 2048          \\
                      &                   &                 & \textbf{s}       & 2.692329                             & 4.774914      & 9.726         & 1024                      & 2048          & 4096          
\end{tblr}
\end{table}